\documentclass{article} 

\usepackage[final]{neurips_2021}

\usepackage[utf8]{inputenc} %
\usepackage[T1]{fontenc}    %
\usepackage{url}            %
\usepackage{subfigure}
\usepackage{booktabs}       %
\usepackage{amsfonts}       %
\usepackage{nicefrac}       %
\usepackage{microtype}      %

\usepackage{wrapfig}

\usepackage[pdftex]{graphicx}
\usepackage{amsmath}
\usepackage[dvipsnames]{xcolor}

\usepackage{algorithmic}
\usepackage{algorithm}
\renewcommand{\algorithmicrequire}{\textbf{Input:}}
\renewcommand{\algorithmicensure}{\textbf{Output:}}

\usepackage{natbib}
\usepackage{enumitem}

\usepackage{xr-hyper}
\usepackage{hyperref}       %

\makeatletter
\newcommand*{\addFileDependency}[1]{%
  \typeout{(#1)}
  \@addtofilelist{#1}
  \IfFileExists{#1}{}{\typeout{No file #1.}}
}
\makeatother

\def\Rset{\mathbb{R}}
\def\cN{\mathcal{N}}
\def\E{\mathbb{E}}

\def\Bset{\mathbb{B}}

\def\bz{\mathbf{z}}
\def\bx{\mathbf{x}}
\def\bs{\mathbf{s}}

\def\vu{\mathbf{u}}

\def\z{\mathbf{z}_t}
\def\y{\mathbf{y}_t}
\def\x{\mathbf{x}_t}

\def\bmu{\boldsymbol{\mu}}

\def\btau{\boldsymbol \tau}
\def\beps{\boldsymbol \epsilon}
\DeclareMathOperator*{\argmax}{\text{arg\,max}}

\def\gN{\mathcal{N}}
\def\gD{\mathcal{D}}
\def\vz{\mathbf{z}}
\def\vx{\mathbf{x}}
\def\vy{\mathbf{y}}

\def\vw{\mathbf{w}}
\def\vmu{\boldsymbol{\mu}}
\def\vtheta{\boldsymbol{\theta}}
\def\veta{\boldsymbol{\eta}}
\def\eps{\epsilon}

\title{Detecting and Adapting to Irregular Distribution Shifts in Bayesian Online Learning}

\author{%
  Aodong Li\textsuperscript{1} \quad Alex Boyd\textsuperscript{2} \quad  Padhraic Smyth\textsuperscript{1,2} \quad Stephan Mandt\textsuperscript{1,2}\\
  \textsuperscript{1}Department of Computer Science \quad \textsuperscript{2}Department of Statistics\\
  University of California, Irvine\\
  \texttt{\{aodongl1, alexjb, mandt\}@uci.edu \quad smyth@ics.uci.edu}
}

\begin{document}

\maketitle

\begin{abstract}

We consider the problem of online learning in the presence of distribution shifts that occur at an unknown rate and of unknown intensity. We derive a new Bayesian online inference approach to simultaneously infer these distribution shifts and adapt the model to the detected changes by integrating ideas from change point detection, switching dynamical systems, and Bayesian online learning. Using a binary ‘change variable,’ we construct an informative prior such that--if a change is detected--the model partially erases the information of past model updates by tempering to facilitate adaptation to the new data distribution. Furthermore, the approach uses beam search to track multiple change-point hypotheses and selects the most probable one in hindsight. Our proposed method is model-agnostic, applicable in both supervised and unsupervised learning settings, suitable for an environment of concept drifts or covariate drifts, and yields improvements over state-of-the-art Bayesian online learning approaches.
\end{abstract}

\section{Introduction}
\label{sec:introduction}

Deployed machine learning systems are often faced with the problem of distribution shift, where the new data that the model processes is systematically different from the data the system was trained on~\citep{zech2018variable,ovadia2019can}. Furthermore, a shift can happen anytime after deployment, unbeknownst to the users, with dramatic consequences for systems such as self-driving cars, robots, and financial trading algorithms, among many other examples. 

Updating a deployed model on new, representative data can help mitigate these issues and improve general performance in most cases.
This task is commonly referred to as \emph{online} or \emph{incremental learning}. Such online learning algorithms face a tradeoff between remembering and adapting. If they adapt too fast, their performance will suffer since adaptation usually implies that the model loses memory of previously encountered training data (which may still be relevant to future predictions). On the other hand, if a model remembers too much, it typically has problems adapting to new data distributions due to its finite capacity. 

The tradeoff between adapting and remembering can be elegantly formalized in a Bayesian online learning framework, where a prior distribution is used to keep track of previously learned parameter estimates and their confidences. For instance, variational continual learning (VCL) \citep{nguyen2017variational} is a popular framework that uses a model's previous posterior distribution as the prior for new data. %
However, the assumption of such continual learning setups is usually that the data distribution is stationary and not subject to change, in which case adaptation is not an issue. 

This paper proposes a new Bayesian online learning framework suitable for non-stationary data distributions. It is based on two assumptions: (i) distribution shifts occur irregularly and must be inferred from the data, and (ii) the model requires not only a good mechanism to aggregate data but also the ability to partially forget information that has become obsolete. 
To solve both problems, we still use a Bayesian framework for online learning (i.e., letting a previous posterior distribution inform the next prior); however, before combining the previously learned posterior with new data evidence, we introduce an intermediate step.
This step allows the model to either broaden the previous posterior's variance to reduce the model's confidence, thus providing more ``room'' for new information, or remain in the same state (i.e., retain the unchanged, last posterior as the new prior).

We propose a mechanism for enabling this decision by introducing a discrete “change variable” that indicates the model's best estimate of whether the data in the new batch is compatible with the previous data distribution or not; the outcome then informs the Bayesian prior at the next time step. We further augment the scheme by performing beam search on the change variable. This way, we are integrating change detection and Bayesian online learning into a common framework. 

We test  our framework on a variety of real-world datasets that show concept drift, including basketball player trajectories, malware characteristics, sensor data, and electricity prices. We also study sequential versions of SVHN and CIFAR-10 with covariate drift, where we simulate the shifts in terms of image rotations. Finally, we study word embedding dynamics in an unsupervised learning approach. Our approach leads to a more compact and interpretable latent structure and significantly improved performance in the supervised experiments. Furthermore, it is highly scalable; we demonstrate it on models with hundreds of thousands of parameters and tens of thousands of feature dimensions.

Our paper is structured as follows: we review related work~in Section~\ref{sec:related}, introduce our methods in Section~\ref{sec:methods}, report our experiments in Section~\ref{sec:experiments}, and draw conclusions in Section~\ref{sec:conclusions}.

\section{Related Work}
\label{sec:related}
Our paper connects to Bayesian online learning, change detection, and switching dynamical systems. 

\paragraph{Bayesian Online and Continual Learning} 
There is a rich existing literature on Bayesian and continual learning. The main challenge in streaming setups is to reduce the impact of old data on the model which can be done by exponentially decaying old samples~\citep{honkela2003line,sato2001online,graepel2010web} or re-weighting them~\citep{mcinerney2015population,theis2015trust}. An alternative approach is to adapt the model posterior between time steps, such as tempering it at a constant rate to accommodate new information \citep{kulhavy1993general,kurle2020continual}. In contrast, \emph{continual learning} typically assumes a stationary data distribution and simply uses the old posterior as the new prior. A scalable such scheme based on variational inference was proposed by \citep{nguyen2017variational} which was extended by several authors \citep{farquhar2019unifying, schwarz2018towards}. A related concept is elastic weight consolidation \citep{kirkpatrick2017overcoming}, where new model parameters are regularized towards old parameter values. 

All of these approaches need to make assumptions on the expected frequency and strengh of change which are hard-coded in the model parameters (e.g., exponential decay rates, re-weighting terms, prior strengths, or temperature choices). Our approach, in contrast, detects change based on a discrete variable and makes no assumption about its frequency.  Other approaches  assume situations where data arrive in irregular time intervals, but are still concerned with static data distributions \citep{titsias2019functional,lee2020neural,rao2019continual}.

\paragraph{Change Point Models} 
There is also a rich history of models for change detection.
A popular class of change point models includes ``product partition models''~\citep{barry1992product} which assume independence of the data distribution across segments. In this regime,
\citet{fearnhead2005exact} proposed change detection in the context of regression and generalized it to online inference~\citep{fearnhead2007line};  
\citet{adams2007bayesian}
described a Bayesian \emph{online} change point detection scheme (BOCD) based on conditional conjugacy assumptions for one-dimensional sequences. Other work generalized change detection algorithms to multivariate time series~\citep{xuan2007modeling, xie2012change} and non-conjugate Bayesian inference~\citep{saatcci2010gaussian,knoblauch2018spatio,turner2013online,knoblauch2018doubly}.

Our approach relies on jointly inferring changes in the data distribution while carrying out Bayesian parameter updates for adaptation. To this end, we detect change in the high-dimensional space of model (e.g., neural network) parameters, as opposed to directly in the data space. Furthermore, a detected change only \emph{partially} resets the model parameters, as opposed to triggering a complete reset. 

\citet{titsias2020sequential} proposed change detection to detect distribution shifts in sequences based on low-dimensional summary statistics such as a loss function; however, the proposed framework does not use an informative prior but requires complete retraining.

\paragraph{Switching Linear Dynamical Systems}
Since our approach integrates a discrete change variable, it is also loosely connected to the topic of  switching linear dynamical systems. 
\citet{linderman2017bayesian} considered \emph{recurrent} switching linear dynamical systems, relying on Bayesian conjugacy and closed-form message passing updates. \citet{becker2019switching} proposed a variational Bayes filtering framework for switching linear dynamical systems. \citet{murphy2012machine} and \citet{barber2012bayesian} developed an inference method using a Gaussian sum filter. Instead, we focus on inferring the full history of discrete latent variable values instead of just the most recent one. 

\citet{bracegirdle2011switch} introduce a \emph{reset} variable that sets the continuous latent variable to an unconditional prior. It is similar to our work, but relies on using low-dimensional, tractable models. Our tempering prior can be seen as a partial reset, augmented with beam search. We also extend the scope of switching dynamical systems by integrating them into a supervised learning framework.

\section{Methods}
\label{sec:methods}
\paragraph{Overview} Section~\ref{sec:model-assumption}
introduces the setup and the novel model structure under consideration. Section~\ref{sec:exact-infer} introduces an exact inference scheme based on beam search. 
Finally, we introduce the variational inference extension for intractable likelihood models in Section~\ref{sec:variational-infer}.

\subsection{Problem Assumptions and Structure}%
\label{sec:model-assumption}
We consider a stream of data that arrives in batches $\x$ at discrete times $t$.\footnote{In an extreme case, it is possible for a batch to include only a single data point.} For supervised setups, we consider pairs of features and targets $(\x,\y)$, where the task is to model $p(\y|\x)$. 
An example model could be a Bayesian neural network, and the parameters $\z$ could be the network weights.
For notational simplicity we focus on the unsupervised case, where the task is to model $p(\x)$ using a model $p(\x|\z)$ with parameters $\z$ that we would like to tune to each new batch.\footnote{
    In supervised setups, we consider a conditional  model $p(\y|\z,\x)$ with features $\x$ and targets $\y$.} 
We then measure the prediction error either on one-step-ahead samples or using a held-out test set.

Furthermore, we assume that while the $\x$ are i.i.d.~within batches, they are not necessarily i.i.d.~across batches as they come from a time-varying distribution $p_t(\x)$ (or $p_t(\x,\y)$ in the supervised cases) which is subject to distribution shifts. We do not assume whether these distribution shifts occur instantaneously or gradually.
The challenge is to optimally adapt the parameters $\z$ to each new batch while borrowing statistical strength from previous batches.%

As follows, we will construct a Bayesian online learning scheme that accounts for changes in the data distribution. For every new batch of data, our scheme tests whether the new batch is compatible with the old data distribution, or more plausible under the assumption of a change. To this end, we employ a binary ``change variable'' $s_t$, with $s_t=0$ for no detected change and $s_t=1$ for a detected change. Our model's joint distribution factorizes as follows:
\begin{equation}
\label{eq:model}
    p(\vx_{1:T}, \vz_{1:T}, s_{1:T}) = \prod_{t=1}^T p(\vx_t|\vz_t) p(\vz_t|s_t; \btau_{t}) p(s_t).
\end{equation}
We assumed a factorized Bernoulli prior $\prod_t p(s_t)$ over the change variable: an assumption that will simplify the inference, but which can be relaxed. As a result, our model is fully-factorized over time, however, the model can still capture temporal dependencies through the informative prior $p(\z | s_t; \btau_t)$. Temporal information enters this prior through certain \emph{sufficient statistics} $\btau_t$ that depend on properties of the previous time-step's approximate posterior. 

In more detail, $\btau_t$ is a \emph{functional} on the previous time step's approximate posterior, $\btau_{t}={\cal F}[p(\vz_{t-1}|\vx_{1:t-1}, s_{1:t-1})]$.\footnote{
    Subscripts $1\!:\!t-1$ indicates the integers from $1$ to $t-1$ inclusively.
}
Throughout this paper, we will use a specific form of $\btau_t$, namely capturing the previous posterior's mean and variance.\footnote{
    In later sections, we will use a Gaussian approximation to the posterior, but here it is enough to assume that these quantities are computable.
    } More formally,
\begin{align}
\label{eq:suff-stat}
\btau_t \equiv \{\bmu_{t-1}, \Sigma_{t-1}\} \equiv \{\text{Mean}, \text{Var}\} [\vz_{t-1}|\vx_{1:t-1}, s_{1:t-1}].
\end{align}
Based on this choice, we define the conditional prior as follows:
\begin{equation}
\label{eq:cond-broaden}
    p(\vz_t|s_t; \btau_{t}) =
    \begin{cases}
        {\cal N}(\vz_t; \bmu_{t-1}, \Sigma_{t-1}) & \textrm{for $s_t=0$} \\ {\cal N}(\vz_t; \bmu_{t-1}, \beta^{-1}\Sigma_{t-1}) & \textrm{for $s_t=1$}
    \end{cases}
\end{equation}
Above, $0 < \beta < 1$ is a hyperparameter referred to as \emph{inverse temperature}\footnote{
   In general it only requires $\beta > 0$ to be inverse temperature. We further assume $\beta<1$ in this paper as this value interval broadens and weakens the previous posterior. See the following paragraphs.
   }. 
If no change is detected (i.e., $s_t=0$), our prior becomes a Gaussian distribution centered around the previous posterior's mean and variance. In particular, if the previous posterior was already Gaussian, it becomes the new prior. In contrast, if a change was detected, the \emph{broadened} posterior becomes the new prior.

For as long as no changes are detected ($s_t=0$), the procedure results in a simple Bayesian online learning procedure, where the posterior uncertainty shrinks with every new observation. In contrast, if a change is detected ($s_t=1$), an overconfident prior would be harmful for learning as the model needs to adapt to the new data distribution. We therefore weaken the prior through \emph{tempering}. Given a temperature $\beta$, we raise the previous posterior's 
Gaussian approximation to the power $\beta$, renormalize it, and then use it as a prior for the current time step.

The process of tempering the Gaussian posterior approximation can be understood as removing equal amounts of information in any direction in the latent space. To see this, let $\vz$ be a multivariate Gaussian  with covariance $\Sigma$ and $\vu$ be a unit direction vector. Then tempering removes an equal amount of information regardless of $\vu$, $H_{\vu} 
= \frac{1}{2}\log(2\pi e \vu^\top \Sigma \vu) - \frac{1}{2}\log \beta$,
erasing learned information to free-up model capacity to adjust to the new data distribution. See Supplement~\ref{sec:app-broadening} for more details.

\paragraph{Connection to Sequence Modeling}
Our model assumptions have a resemblance to time series modeling: if we replaced $\btau_{t}$ with $\vz_{t-1}$, we would condition on previous latent states rather than posterior summary statistics. 
In contrast, our model still factorizes over time and therefore makes weaker assumptions on temporal continuity. 
Rather than imposing temporal continuity on a data \emph{instance} level, we instead assume temporal continuity at a \emph{distribution} level. 

\paragraph{Connection to Changepoint Modeling.}
We also share similar assumptions with the changepoint literature~\citep{barry1992product}. However, in most cases, these models don’t assume an informative prior, effectively not taking into account any sufficient statistics $\btau_t$. 
This forces these models to re-learn model parameters from scratch after a detected change, whereas our approach allows for some transfer of information before and after the distribution shift.

\subsection{Exact Inference}
\label{sec:exact-infer}

\begin{figure}
    \begin{center}
        \includegraphics[width=0.45\columnwidth]{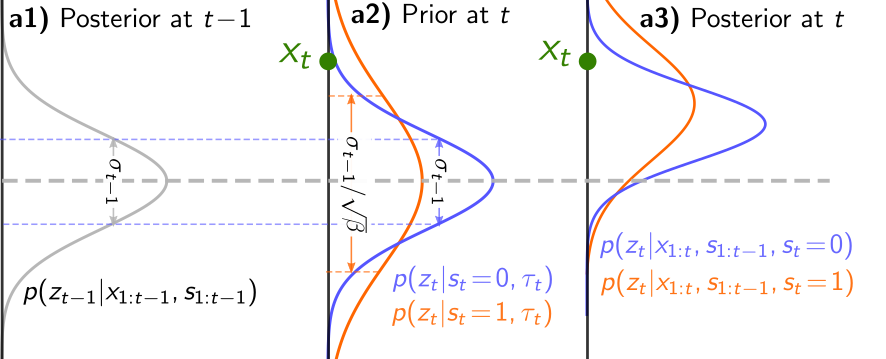}
        \includegraphics[width=0.5\columnwidth]{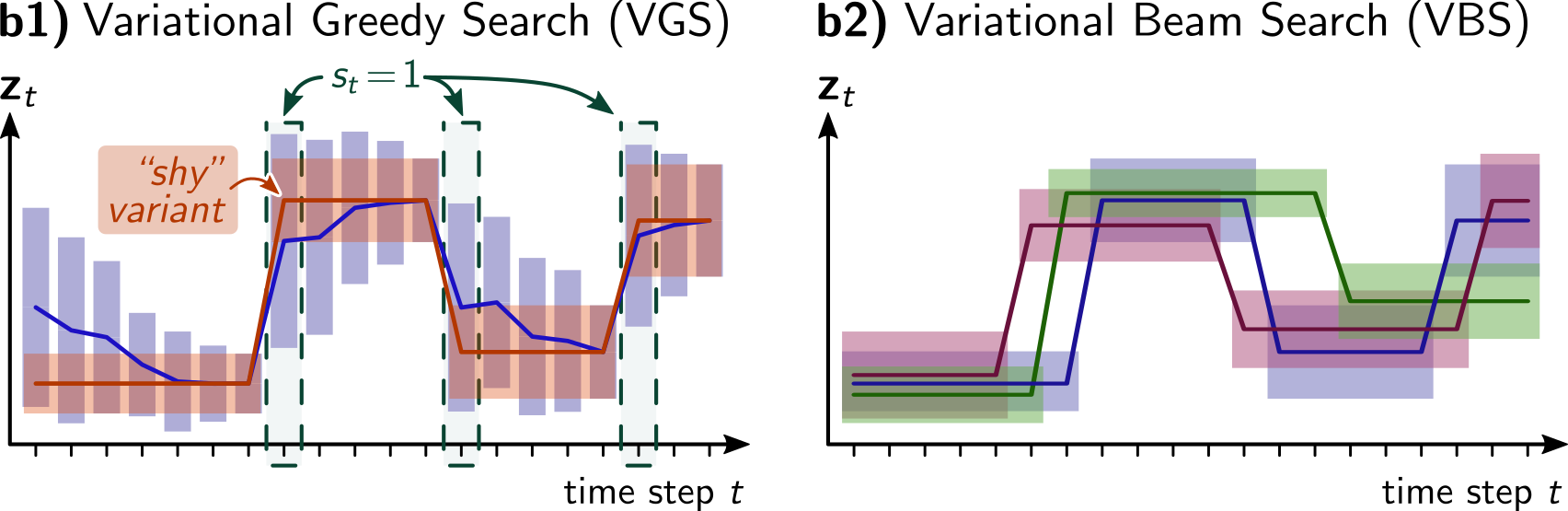}
    \caption{\textbf{a)} A single inference step for the latent mean in a 1D linear Gaussian model. Starting from the previous posterior (\textbf{a1}), we consider both its broadened and un-broadened version (\textbf{a2}). Then the model absorbs the observation and updates the priors (\textbf{a3}).
    \textbf{b)} Sparse inference via greedy search (\textbf{b1}) and variational beam search (\textbf{b2}). b) Solid lines indicate fitted mean $\mu_t$ over time steps $t$ with boxes representing $\pm 1\sigma$ error bars. See more information about the pictured ``shy'' variant in Supplement~\ref{sec:app-beamsearch}.}
    \label{fig:filtering}
    \end{center}
\end{figure}

Before presenting a scalable variational inference scheme in our model, we describe an exact inference scheme when everything is tractable, i.e., the special case of linear Gaussian models. 

According to our assumptions, the distribution shifts occur at discrete times and are unobserved. 
Therefore, we have to infer them from the observed data and adapt the model accordingly. Recall the distribution shift is represented by the binary latent variable $s_t$ at time step $t$. Inferring the posterior over $s_t$ at $t$ will thus notify us how likely the change happens under the model assumption. As follows, we show the posterior of $s_t$ is simple in a tractable model and bears similarity with a likelihood ratio test. Suppose we moved from time step $t-1$ to step $t$ and observed new data $\vx_t$. Denote the history decisions and observations by $\{s_{1:t-1}, \vx_{1:t-1}\}$, which enters through $\btau_t$. Then by Bayes rule, the exact posterior over $s_t$ is again a Bernoulli, $p(s_t|s_{1:t-1}, \vx_{1:t}) = {\rm Bern}(s_t; m)$, with parameter
\begin{equation}
\label{eq:exact-inference}
\begin{aligned}
m = \sigma\left(\log \frac{p(\x|s_t\!=\!1, s_{1:t-1}, \vx_{1:t-1}) p(s_t\!=\!1)}{p(\x|s_t\!=\!0, s_{1:t-1}, \vx_{1:t-1})p(s_t\!=\!0)}\right) 
             = \sigma\left(\log \frac{p(\x|s_t\!=\!1, s_{1:t-1}, \vx_{1:t-1})}{p(\x|s_t\!=\!0, s_{1:t-1}, \vx_{1:t-1})} + \xi_0\right).
\end{aligned}
\end{equation}
Above, $\sigma$ is the sigmoid function, and $\xi_0\!=\!\log p(s_t\!=\!1) - \log p(s_t\!=\!0)$ are the log-odds of the prior $p(s_t)$ and serves as a bias term. $p(\x|s_{1:t}, \vx_{1:t-1})=\int p(\x|\z) p(\z|s_t; \btau_{t}) d\z$ is the model evidence. Overall, $m$ specifies the probability of $s_t=1$ given $\vx_{1:t}$ and $s_{1:t-1}$.

Eq.~\ref{eq:exact-inference} has a simple interpretation as a likelihood ratio test: a change is more or less likely depending on whether or not the observations $\x$ are better explained under the assumption of a detected change.

We have described the detection procedure thus far, now we turn to the adaptation procedure. To adjust the model parameters $\vz_t$ to the new data given a change or not, we combine the likelihood of $\vx_t$ with the conditional prior (Eq.~\ref{eq:cond-broaden}). This corresponds to the posterior of $\vz_t$, $p(\vz_t|\vx_{1:t}, s_{1:t}) = \frac{p(\vx_t|\vz_t)p(\vz_t|s_t; \btau_{t})}{p(\x|s_{1:t}, \vx_{1:t-1})}$, obtained by Bayes rule.
The adaptation procedure is illustrated in Fig.~\ref{fig:filtering} (a), where we show how a new observation modifies the conditional prior of model parameters.

As a result of Eq.~\ref{eq:exact-inference}, we obtain the marginal distribution of $\vz_t$ at time $t$ as a binary mixture with mixture weights $p(s_t\!=\!1|s_{1:t-1}, \vx_{1:t})=m$ and $p(s_t\!=\!0|s_{1:t-1}, \vx_{1:t})=1-m$: $p(\vz_t|s_{1:t-1}, \vx_{1:t})=m p(\vz_t|s_{t}\!=\!1, s_{1:t-1}, \vx_{1:t}) + (1\!-\!m) p(\vz_t|s_{t}\!=\!0, s_{1:t-1}, \vx_{1:t})$.

\paragraph{Exponential Branching} We note that while we had originally started with a posterior $p(\vz_{t-1}|\vx_{1:t-1}, s_{1:t-1})$ at the previous time, our inference scheme resulted in $p(\vz_t|s_{1:t-1}, \vx_{1:t})$ being a mixture of two components as it branches over two possible states.\footnote{
See also Fig.~\ref{fig:beam-search-table} in Supplement~\ref{sec:app-beamsearch}.} When we iterate, we encounter an exponential branching of possibilities, or \emph{hypotheses} over possible sequences of regime shifts $s_{1:t}$. To still carry out the filtering scheme efficiently, we need a truncation scheme, e.g., approximate the bimodal marginal distribution by a unimodal one. As follows, we will discuss two methods---greedy search and beam search---to achieve this goal.

\paragraph{Greedy Search}
In the simplest ``greedy'' setup, we train the model in an online fashion by iterating over time steps $t$. For each $t$, we update a \emph{truncated} distribution via the following three steps:

\begin{enumerate}[leftmargin=4.5mm]
\item Compute the conditional prior $p(\vz_t|s_t; \btau_{t})$ (Eq.~\ref{eq:cond-broaden}) based on $p(\vz_{t-1}|\vx_{1:t-1}, s_{1:t-1})$ and evaluate the likelihood $p(\vx_t|\vz_t)$ upon observing data $\vx_t$.
\item Infer whether a change happens or not using the posterior over $s_t$ (Eq.~\ref{eq:exact-inference}) and adapt the model parameters $\vz_t$ for each case. 
\item Select $s_{t}\in\{0,1\}$ that has larger posterior probability $p(s_t|s_{1:t-1}, \vx_{1:t})$ and its corresponding model hypothesis $p(\vz_t|s_{1:t}, \vx_{1:t})$ (i.e., make a ``hard'' decision over $s_t$ with a threshold of $\frac{1}{2}$).
\end{enumerate}

The above filtering algorithm iteratively updates the posterior distribution over $\z$ each time it observes new data $\x$. In the version of greedy search discussed above, the approach decides immediately, i.e., before observing subsequent data points, whether a change in $\z$ has occurred or not in step 3. (Please note the decision is irrelevant to history, as opposed to the beam search described below.) 
We illustrate greedy search is illustrated in Fig.~\ref{fig:filtering}~(b1) where VGS is the variational inference counterpart.

\paragraph{Beam Search} A greedy search is prone to missing change points in data sets with a low signal/noise ratio per time step because it cannot accumulate evidence for a change point over a series of time steps.  
The most obvious improvement over greedy search that has the ability to accumulate evidence for a change point is beam search. Rather than deciding greedily whether a change occurred or not at each time step, beam search considers both cases in parallel as it delays the decision of which one is more likely (see Fig.~\ref{fig:filtering}~(b2) and Fig.~\ref{fig:toydata}~(left) for illustration). The algorithm keeps track of a fixed number $K>1$ of possible hypotheses of change points. For each hypothesis, it iteratively updates the posterior distribution as a greedy search. At time step $t$, every potential continuation of the $K$ sequences is considered with $s_t\in\{0,1\}$, thus doubling the number of histories of which the algorithm has to track. To keep the computational requirements bounded, beam search thus discards half of the sequences based on an exploration-exploitation trade-off.

Beam search simultaneously tracks multiple hypotheses necessitating the differentiation between them. In greedy search, we can distinguish hypotheses based on the most recent $s_t$'s value since only two hypotheses are considered at each step. However, beam search considers at most $2K$ hypotheses each step, which exceeds the capacity of a single $s_t$. We thus resort to the decision history $s_{1:t-1}$ to further tell hypotheses apart. The weight $p(s_{1:t}|\vx_{1:t})$ of each hypothesis can be computed recursively:
\begin{align}
    p(s_{1:t}|\vx_{1:t})&~\propto p(s_t, \vx_t|s_{1:t-1}, \vx_{1:t-1})p(s_{1:t-1}|\vx_{1:t-1}) \nonumber
    \\&~\propto p(s_t|s_{1:t-1}, \vx_{1:t})p(s_{1:t-1}|\vx_{1:t-1})\label{eq:s-recursion}
\end{align} 
where the added information $p(s_t|s_{1:t-1}, \vx_{1:t})$ at step $t$ is the posterior of $s_t$ (Eq.~\ref{eq:exact-inference}). This suggests the ``correction in hindsight'' nature of beam search: re-ranking the sequence $s_{1:t}$ as a whole at time $t$ indicates the ability to correct decisions before time $t$. 

Another ingredient is a set ${\mathbb B}_t$, which contains the $K$ most probable “histories” $s_{1:t}$ at time $t$. From time $t-1$ to $t$, we evaluate the continuation of each hypothesis $s_{1:t-1}\in{\mathbb B}_{t-1}$ as the first two steps of greedy search, leading to $2K$ hypotheses. We then compute the weight of each hypothesis using Eq.~\ref{eq:s-recursion}. Finally, select top $K$ hypotheses into ${\mathbb B}_t$ and re-normalize the weights of hypotheses in ${\mathbb B}_t$.

This concludes the recursion from time $t-1$ to $t$. With $p(\vz_{t}|s_{1:t}, \vx_{1:t})$ and $p(s_{1:t}|\vx_{1:t})$, we can achieve any marginal distribution of $\vz_t$, such as $p(\vz_t|\vx_{1:t})=\sum_{s_{1:t}}p(\vz_{t}|s_{1:t}, \vx_{1:t})p(s_{1:t}|\vx_{1:t})$. %

\paragraph{Beam Search Diversification}
Empirically, we find that the naive beam search procedure does not realize its full potential. As commonly encountered in beam search, histories over change points are largely shared among all members of the beam. To encourage diverse beams, we constructed the following simple scheme. While transitioning from time $t\!-\!1$ to $t$, every hypothesis splits into two scenarios, one with $s_t\!=\!0$ and one with $s_t\!=\!1$, resulting in $2K$ temporary hypotheses. If two resulting hypotheses only differ in their most recent $s_t$-value, we say that they come from the same ``family.''  Each member among the $2K$ hypotheses is ranked according to its posterior probability $p(s_{1:t}|\vx_{1:t})$ in Eq.~\ref{eq:s-recursion}. In a first step, we discard the bottom $1/3$ of the $2K$ hypotheses, leaving  $4/3 K$ hypotheses (we always take integer multiples of $3$ for $K$). To truncate the beam size from $4/3K$ down to $K$, we rank the remaining hypotheses according to their posterior probability and pick the top $K$ ones while \emph{also} ensuring that we pick a member from every remaining family. The diversification scheme ensures that underperforming families can survive, leading to a more diverse set of hypotheses. We found this beam diversification scheme to work robustly across a variety of experiments.

\subsection{Variational Inference}
\label{sec:variational-infer}
In most practical applications, the evidence term is not available in closed-form, leaving Eq.~\ref{eq:exact-inference} intractable to evaluate. However, we can follow a structured variational inference approach~\citep{wainwright2008graphical,hoffman2015structured,zhang2018advances}, defining a joint variational distribution $q(\z,s_t|s_{1:t-1}) = q(s_t|s_{1:t-1})q(\z|s_{1:t})$, to approximate $p(\z, s_t|s_{1:t-1}, \vx_{1:t})=p(s_t|s_{1:t-1}, \vx_{1:t})p(\z|s_{1:t}, \vx_{1:t})$. This procedure completes the detection and adaptation altogether. 

One may wonder how the exact inference schemes for $s_t$ and $\vz_t$ are modified in the structured variational inference scenario. In Supplement~\ref{sec:app-structvi}, we derive the solution for $q(\vz_t, s_t|s_{1:t-1})$. Surprisingly we have the following closed-form update equation for $q(s_t|s_{1:t-1})$ that bears strong similarities to Eq.~\ref{eq:exact-inference}. The new scheme simply replaces the intractable evidence term with a lower bound proxy -- optimized conditional evidence lower bound ${\cal L}(q^*| s_{1:t})$ (CELBO, defined later), giving the update
\begin{equation}
\label{eq:update_s}
    q^*(s_t|s_{1:t-1}) = {\rm Bern}(s_t; m); \quad m = \textstyle \sigma\left(\frac{1}{T}{\cal L}(q^*| s_t\!=\!1, s_{1:t-1}) - \frac{1}{T}{\cal L}(q^*| s_t\!=\!0, s_{1:t-1}) + \xi_0\right).
\end{equation}
Above, we introduced a parameter $T\geq 1$ (not to be confused with $\beta$) to optionally downweigh the data evidence relative to the prior (see Experiments Section~\ref{sec:experiments}).  

Now we define the CELBO. To approximate $p(\vz_t|s_{1:t}, \vx_{1:t})$ by variational distribution $q(\z|s_{1:t})$,
we minimize the KL divergence between $q(\z|s_{1:t})$ and $p(\vz_t|s_{1:t}, \vx_{1:t})$,
leading to
\begin{align}
\label{eq:cond-elbo}
    & q^*(\z|s_{1:t})    =  \argmax_{q(\z|s_{1:t}) \in Q} {\cal L}(q|s_{1:t}), \\
    {\cal L}(q|s_{1:t})& := \E_q [\log p(\x|\z)] - \textrm{KL}(q(\z|s_{1:t})||p(\vz_t|s_t; \btau_{t})). \nonumber
\end{align}
$Q$ denotes the variational family (i.e., factorized normal distributions), and we term ${\cal L}(q|s_{1:t})$ CELBO.

The greedy search and beam search schemes also apply to variational inference. We name them \textit{variational greedy search} (VGS, VBS (K=1)) and \textit{variational beam search} (VBS) (Fig.~\ref{fig:filtering} (b)). 

\paragraph{Algorithm Complexity}
VBS’s computational time and space complexity scale \textit{linearly} with the beam size $K$. As such, its computational cost is only about $2K$ times larger than greedy search\footnote{This also applies to the baselines  ``Bayesian Forgetting'' (BF) and 
Variational Continual Learning" (VCL) introduced in Section~\ref{sec:exp-baseline}.}. 
Furthermore, our algorithm’s complexity is $O(1)$ in the sequence length $t$. It is not necessary to store sequences $s_{1:t}$ as they are just symbols to distinguish hypotheses.
The only exception to this scaling would be an application asking for the most likely changepoint sequence in hindsight. In this case, the changepoint sequence (but not the associated model parameters) would need to be stored, incurring a cost of storing exactly $K \times T$ binary variables. This storage is, however, not necessary when the focus is only on adapting to distribution shifts.

\section{Experiments}
\label{sec:experiments}
\paragraph{Overview} 
The objective of our experiments is to show that, compared to other methods, variational beam search  (1)  better reacts to different distribution shifts, e.g., \textit{concept drifts} and \textit{covariate drifts},
while (2) revealing interpretable and temporally sparse latent structure. We experiment on artificial data to demonstrate the “correct in hindsight” nature of VBS (Section~\ref{sec:exp-toydata}),  evaluate online linear regression on three datasets with concept shifts (Section~\ref{sec:exp-linreg}), visualize the detected change points on basketball player movements, demonstrate the robustness of the hyperparameter $\beta$ (Section~\ref{sec:exp-linreg}), study Bayesian deep learning approaches on sequences of transformed images with covariate shifts (Section \ref{sec:exp-bdl}), and study the dynamics of word embeddings on historical text corpora (Section~\ref{sec:exp-embeddings}). 
Unstated experimental details are in Supplement~\ref{sec:app-expdetails}.

\begin{figure}
    \centering
     \includegraphics[width=0.33\columnwidth]{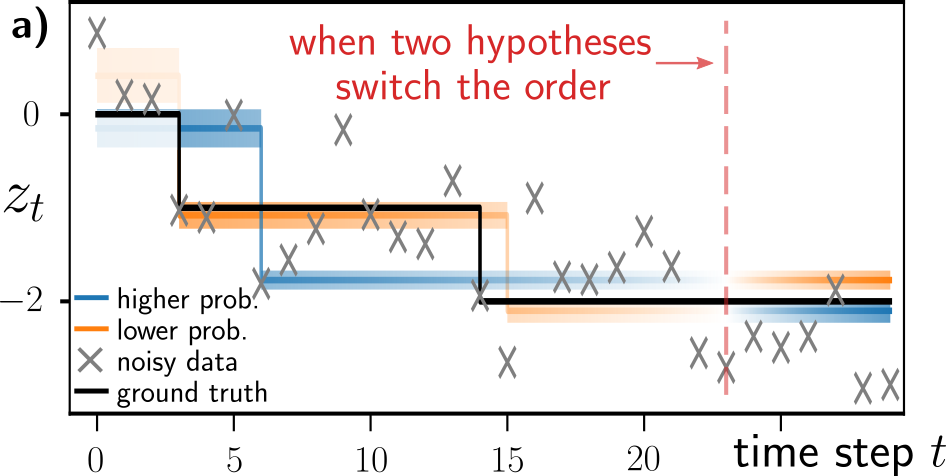}
     \includegraphics[width=0.3\columnwidth]{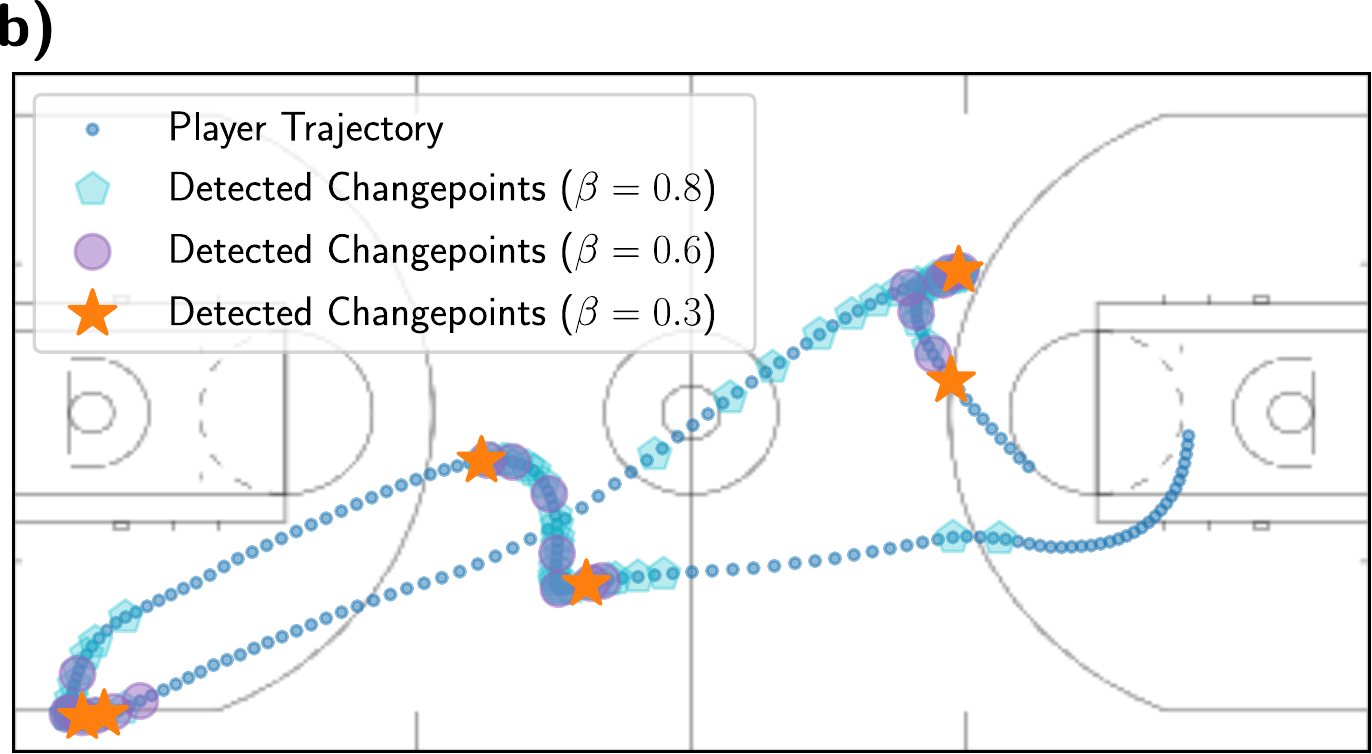}
    \includegraphics[width=0.33\columnwidth]{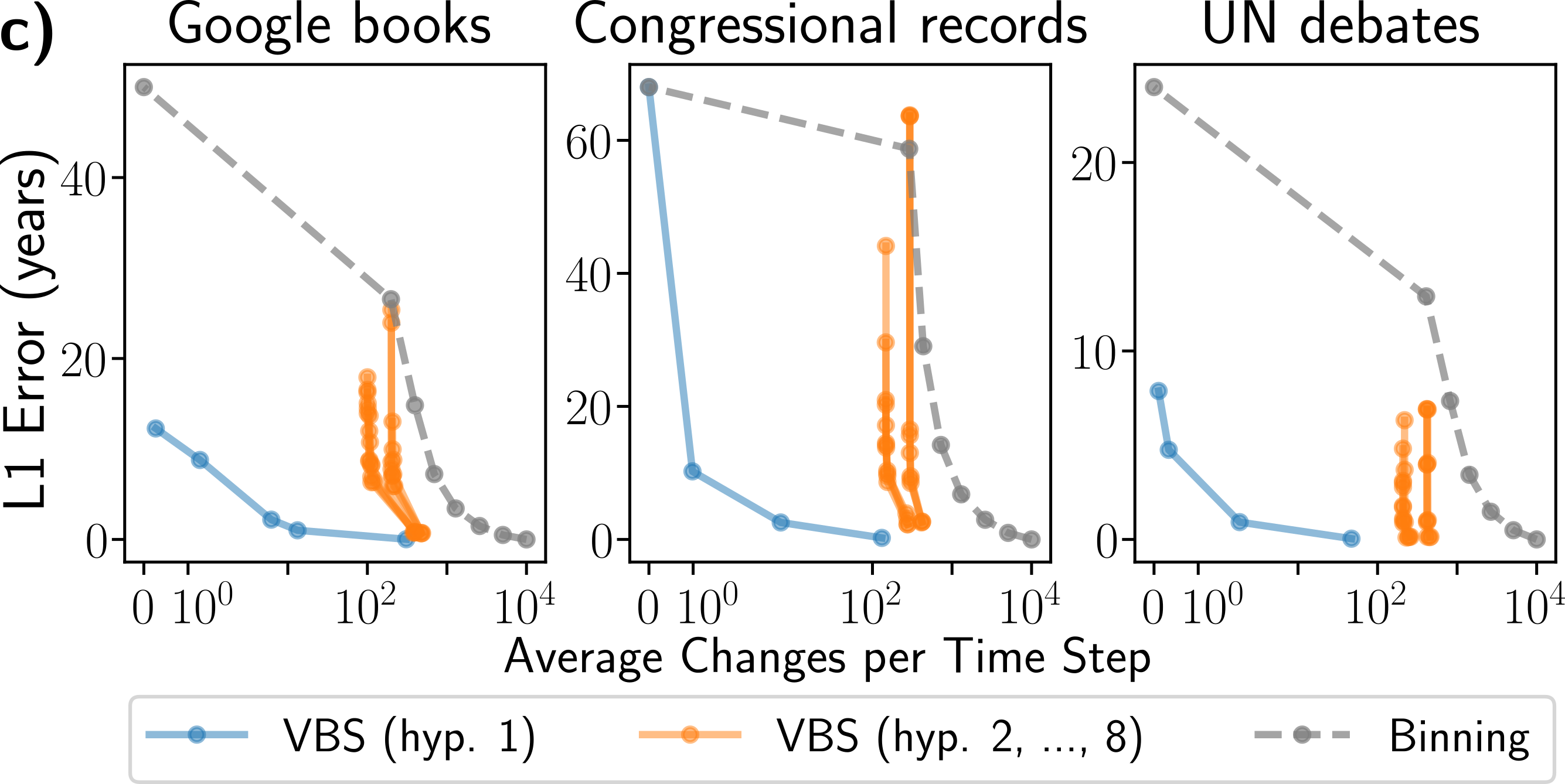}
     
     \caption{\textbf{a)} Inferring the mean (black line) of a time-varying data distribution (black samples) with VBS. The initially unlikely hypothesis begins dominating over the other at step 23. \textbf{b)} Basketball player tracking: ablation study over $\beta$ for VBS while fixing other parameters. We used   greedy search (K=1) and run the model under different $\beta$ values. Increasing $\beta$ leads to more sensitivity to changes in data, leading to more detected changepoints. 
     \textbf{c)} Document dating error as a function of model sparsity, measured in average words update per year. As semantic changes get successively sparsified by varying $\xi_0$ (Eq.~\ref{eq:update_s}), VBS maintains a better document dating performance compared to baselines.
     }
    \label{fig:toydata}
\end{figure}

\subsection{An Illustrative Example}\label{sec:exp-toydata}

We first aim to dynamically demonstrate the ``correction in hindsight'' nature of VBS based on a simple setup involving artificial data.
To this end, we formulated a problem of tracking the shifting mean of data samples. This shifting mean is a piecewise-constant step function involving two steps (seen as in black in Fig.~\ref{fig:toydata} (a)), and we simulated noisy data points centered around the mean. We then used VBS with beam size $K=2$ to fit the same latent mean model that generated the data. The color indicates the ranking among both hypotheses at each moment in time (blue as ``more likely'' vs. orange as ``less likely''). While hypothesis 1 assumes a single distribution shift (initially blue), hypothesis 2 (initially orange) assumes two shifts. We see that hypothesis 1 is initially more likely, but gets over-ruled by the better hypothesis 2 later (note the color swap at step 23).

\subsection{Baselines}\label{sec:exp-baseline}
In our supervised experiments (Section~\ref{sec:exp-linreg} and Section~\ref{sec:exp-bdl}), we compared VBS against adaptive methods, Bayesian online learning baselines, and independent batch learning baselines.\footnote{
  As a reminder, a “batch” at discrete time $t$ is the dataset available for learning; on the other hand, a “mini-batch” is a small set of data used for computing gradients for stochastic gradient-based optimization.
} 
Among the adaptive methods, we formulated a supervised learning version of Bayesian online changepoint detection (BOCD)~\citep{adams2007bayesian}.\footnote{
  1) Although BOCD is mostly applied for unsupervised learning, its application in supervised learning and its underlying model's adaptation to change points are seldom investigated. 2) When the model is non-conjugate, such as Bayesian neural networks, we approximate the log evidence $\log p(y|x)$ by the evidence lower bound.
} 
We also implemented Bayesian forgetting (BF)~\citep{kurle2020continual} with convolutional neural networks for proper comparisons. Bayesian online learning baselines include variational continual learning (VCL) ~\citep{nguyen2017variational} and Laplace propagation (LP)~\citep{smola2003laplace, nguyen2017variational}. 
Finally, we also adopt a trivial baseline of learning independent regressors/classifiers on each batch in both a Bayesian and non-Bayesian fashion. 
For VBS and BOCD we always report the most dominant hypothesis.
In unsupervised learning experiments, we compared against the online version of word2vec~\citep{mikolov2013distributed} with a diffusion prior, dynamic word embeddings~\citep{bamler2017dynamic}.

\subsection{Bayesian Linear Regression Experiments}\label{sec:exp-linreg}
As a simple first version of VBS, we tested an online linear regression setup for which the posterior can be computed analytically.  The analytical solution removes the approximation error of the variational inference procedure as well as optimization-related artifacts since closed-form updates are available. 
Detailed derivations are in Supplement~\ref{sec:app-linreg}.

\begin{figure}%
\begin{minipage}[b]{0.65\textwidth}
\begin{table}[H]
\caption{Evaluation of Different Datasets} \label{table-bdlaccuracy}
\vskip 0.1in
\begin{center}
\begin{tiny}
\begin{sc}
\begin{tabular}{@{\hskip 2pt} l @{\hskip 4pt} c @{\hskip 4pt} c @{\hskip 4pt} c @{\hskip 4pt} c @{\hskip 4pt} c @{\hskip 2pt} c @{\hskip 2pt}}
\toprule
Models &CIFAR-10 &SVHN &Malware &SensorDrift &Elec2 &NBAPlayer \\
&\multicolumn{2}{c}{(Accuracy)$\uparrow$} &\multicolumn{3}{c}{(MCAE $10^{\!-\!2}$)$\downarrow$} &\multicolumn{1}{c @{\hskip 0pt}}{(LogLike $10^{\!-\!2}$)$\uparrow$} \\
\midrule
VBS (K=6)$^*$ &\textbf{69.2$\pm$0.9} &\textbf{89.6$\pm$0.5} &\textbf{11.61} &\textbf{10.53} &7.28 &\textbf{29.49$\pm$3.12} \\
VBS (K=3)$^*$ &68.9$\pm$0.9 &89.1$\pm$0.5  &11.65 &10.71 &7.28 & \textbf{29.22$\pm$2.63}\\
VBS (K=1)$^*$ &68.2$\pm$0.8 &88.9$\pm$0.5  &11.65 &10.86 &\textbf{7.27} &\textbf{29.25$\pm$2.59}\\
\cmidrule{1-7}%
BOCD (K=6)$^\sharp$ &65.6$\pm$0.8 &88.2$\pm$0.5  &12.93 &24.34 &12.49 &22.96$\pm$7.42\\
BOCD (K=3)$^\sharp$ &67.3$\pm$0.8 &88.8$\pm$0.5  &12.74 &24.31 &12.49 &20.93$\pm$7.83\\
BF$^\P$ &\textbf{69.8$\pm$0.8} &\textbf{89.9$\pm$0.5}  &11.71 &11.40 &13.37 &24.17$\pm$2.29\\
VCL$^\dagger$ &66.7$\pm$0.8 &88.7$\pm$0.5  &13.27 &24.90 &16.59 &3.48$\pm$25.53\\
LP$^\ddagger$ &62.6$\pm$1.0 &82.8$\pm$0.9 &13.27 &24.90 &16.59 &3.48$\pm$25.53\\
IB$^\S$ &63.7$\pm$0.5 &85.5$\pm$0.7 &16.6 &27.71 &12.48 &-44.87$\pm$16.88\\
IB$^\S$ (Bayes) &64.5$\pm$0.3 &87.8$\pm$0.1 &16.6 &27.71 &12.48 &-44.87$\pm$16.88\\
\bottomrule
\multicolumn{7}{@{\hskip 0pt} l @{\hskip 0pt}}{{\tiny $^*$ proposed, $^\sharp$ [Adams and MacKay, 2007], $^\P$ [Kurle et al., 2020]}}  \\
\multicolumn{7}{@{\hskip 0pt} l @{\hskip 0pt}}{{\tiny $^\dagger$ [Nguyen et al., 2018], $^\ddagger$ [Smola et al., 2003], $^\S$ Independent Batch}}
\end{tabular}
\end{sc}
\end{tiny}
\end{center}
\vskip -0.1in
\end{table}

\end{minipage}
\begin{minipage}[b]{0.33\textwidth}
\begin{figure}[H]
     \center
     \includegraphics[width=0.9\textwidth]{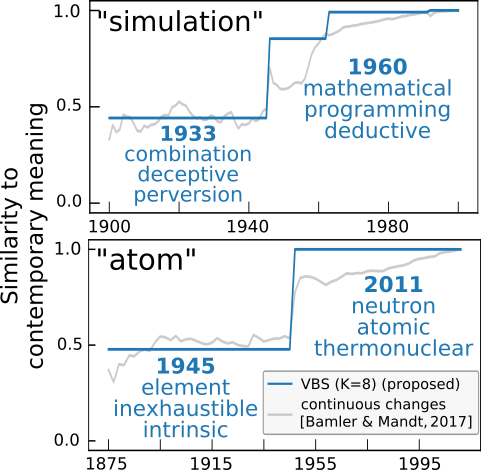}
     \caption{\small Sparse word meaning changes in ``simulation'' and ``atom''.
     }
     \label{fig:dwe}
\end{figure}
\end{minipage}

\end{figure}

\paragraph{Real Datasets with Concept Shifts.} We investigated three real-world datasets with \textit{concept shifts}:
\begin{itemize}[leftmargin=4.5mm]
    \item \textbf{Malware} This dataset is a collection of 100K malignous and benign computer programs, collected over 44 months \citep{huynh2017new}. Each program has 482 counting features and a real-valued probability $p\in[0,1]$ of being malware. We linearly predicted the log-odds. 
    \item \textbf{SensorDrift} A collection of chemical sensor readings \citep{vergara2012chemical}. We predicted the concentration level of gas \emph{acetaldehyde}, whose 2,926 samples and 128 features span 36 months. %
    \item \textbf{Elec2} The dataset contains the electricity price over three years of two Australian states \citep{harries1999splice}. While the original problem formulation used a majority vote to generate 0-1 binary labels on whether the price increases or not, we averaged the votes out into real-valued probabilities and predicted the log-odds instead. We had 45,263 samples and 14 features. 
\end{itemize}
At each step, only one data sample is revealed to the regressor. We evaluated all methods with one-step-ahead absolute error\footnote{We measured the error in the probability space for classification problems (Malware and Elec2) and the error in the data space for regression problems (SensorDrift).} and computed the mean cumulative absolute error (MCAE) at every step. In Table~\ref{table-bdlaccuracy}, we didn't report the variance of MCAEs since there is no stochastic optimization noise. Table~\ref{table-bdlaccuracy} shows that VBS has the best average of MCAEs among all methods. 
We also reported the running performance in Supplement~\ref{sec:app-linregexp}, where other experimental details are available as well.

\paragraph{Basketball Player Tracking.} We explored a collection of basketball player movement trajectories.\footnote{
    \url{https://github.com/linouk23/NBA-Player-Movements}
    }
Each trajectory has wide variations in player velocities. We treated the trajectories as time series and used a Bayesian transition matrix to predict the next position $\vx_{t+1}$ based on the current position $\vx_t$. This matrix is learned and adapted on the fly for each trajectory. 

We first investigated the effect of the temperature parameter $\beta$ in our approach. To this end, we visualized the detected change points on an example trajectory. We used VBS (K=1, greedy search) and compared different values of $\beta$ in Fig.~\ref{fig:toydata} (b). The figure shows that the larger $\beta$, the more change points are detected; the smaller $\beta$, the detected change points get sparser, i.e., $\beta$ determines the model's sensitivity to changes. This observation confirms the assumption that $\beta$ controls the assumed strength of distribution shifts. 

In addition, the result also implies the robustness of poorly selected $\beta$s. When facing an abrupt change in the trajectory, the regressor has two adapt options based on different $\beta$s -- make a single strong adaptation or make a sequence of weak adaptations -- in either case, the model ends up adapting itself to the new trajectory. In other words, people can choose different $\beta$ for a specific environment, with a trade-off between adaptation speed and the erased amount of information.

Finally, regarding the quantitative results, we evaluated all methods with the time-averaged predictive log-likelihood on a reserved test set in Table~\ref{table-bdlaccuracy}. Our proposed methods yield better performance than the baselines. In Supplement~\ref{sec:app-nbaplayer}, we provide more results of change point detection.

\subsection{Bayesian Deep Learning Experiments}\label{sec:exp-bdl}

Our larger-scale experiments involve Bayesian convolutional neural networks trained on sequential batches for image classification using  CIFAR-10~\citep{krizhevsky2009learning} and SVHN~\citep{netzer2011reading}. Every few batches, we manually introduce \textit{covariate shifts} through transforming all images globally by combining rotations, shifts, and scalings.  Each transformation is generated from a fixed, predefined distribution (see Supplement~\ref{sec:app-bdl}). The experiment involved 100 batches in sequence, where each batch contained a third of the transformed datasets. We set the temperature $\beta\!=\!2/3$ and set the CELBO temperature $T\!=\!20,000$ (in Eq.~\ref{eq:update_s}) for all supervised experiments. 

Table~\ref{table-bdlaccuracy} shows the performances of all considered methods and both data sets, averaged across all of the 100 batches. Within their confidence bounds, VBS and BF have comparable performances and outperform the other baselines. We conjecture that the strong performance of BF can be attributed to the fact that our imposed changes are still relatively evenly spaced and regular. 
The benefit of beam search in VBS is evident, with larger beam sizes consistently performing better.

\subsection{Unsupervised Experiments}\label{sec:exp-embeddings}

Our final experiment focused on unsupervised learning. We intended to demonstrate that VBS helps uncover interpretable latent structure in high-dimensional time series by detecting change points.
We also showed that the detected change points help reduce the storage size and maintain salient features.

Towards this end, we analyzed the semantic changes of individual words over time in an unsupervised setup. We used Dynamic Word Embeddings (DWE)~\citep{bamler2017dynamic} as our base model. The model is an online version of Word2Vec~\citep{mikolov2013distributed}. Word2Vec projects a vocabulary into an embedding space and measures word similarities by cosine distance in that space. DWE further imposes a time-series prior on the embeddings and tracks them over time.
For our proposed approach, we augmented DWE with VBS, allowing us to detect the changes of words meaning.

We analyzed three large time-stamped text corpora, all of which are available online. Our first dataset is the Google Books corpus~\citep{michel2011quantitative} in $n$-grams form.
We focused on 1900 to 2000 with sub-sampled 250M to 300M tokens per year. 
Second, we used the Congressional Records dataset~\citep{gentzkow2018congressional}, which has 13M to 52M tokens per two-year period from 1875 to 2011. Third, we used the UN General Debates corpus~\citep{DVN/0TJX8Y_2017}, which has about 250k to 450k tokens per year from 1970 to 2018.

Our first experiments demonstrate VBS provides more interpretable step-wise word meaning shifts than the continuous shifts (DWE). Due to page limits, in Fig.~\ref{fig:dwe} we selected two example words and their three nearest neighbors in the embedding space at different years. The evolving nearest neighbors reflect a semantic change of the words. We plotted the most likely hypothesis of VBS in blue and the continuous-path baseline (DWE) in grey. 
While people can roughly tell the change points from the continuous path, the changes are surrounded by noisy perturbations and sometimes submerged within the noise. VBS, on the other hand, makes clear decisions  and outputs explicit change points. As a result, VBS discovers that the word “atom” changes its meaning  from ``element'' to ``nuclear'' in 1945--the year when two nuclear bombs were detonated; 
word ``simulation'' changes its dominant context from ``deception'' to ``programming'' with the advent of computers in the 1950s.
Besides interpretable changes points, VBS provides multiple plausible hypotheses (Supplement~\ref{sec:app-embeddings}). 

Our second experiments exemplify the usefulness of the detected \textit{sparse} change points, which lead to sparse segments of embeddings. The usefulness comes in two folds: 1) while alleviating the burden of the disk storage by storing one value for each segment, 2) the sparsity preserves the salient features of the original model. To illustrate these two aspects, we design a document dating task that exploits the probabilistic interpretation of word embeddings. The idea is to assign a test document to the year whose embeddings provide the highest likelihood. In Figure~\ref{fig:toydata} (c), we measure the model sparsity on the x-axis with the average updated embeddings per step (The maximum is 10000, which is the vocabulary size). The feature preservation ability is measured by document dating accuracy on the y-axis. We adjust the prior log-odds $\xi_0$ (Eq.~\ref{eq:update_s}) to have successive models with different change point sparsity and then measure the dating accuracy. We also designed an oracle baseline named “binning” (grey, Supplement~\ref{sec:app-embeddings}). For VBS, we show the dominant hypothesis (blue) as well as the subleading hypotheses (orange). The most likely hypothesis of VBS outperforms the baseline, leading to higher document dating precision at much smaller disk storage.

\section{Discussion}\label{sec:method-discuss}

\paragraph{Beyond Gaussian Posterior Approximations.}
While the Gaussian approximation is simple and is widely used (and broadly effective) in practice in Bayesian inference [e.g., \citet{murphy2012machine}, pp.649-662], our formulation does not rule out the extensions to exponential families. $\btau_t$ in Eq.~\ref{eq:suff-stat} could be generalized by reading off sufficient statistics of the previous approximate posterior. To this end, we need a sufficient statistic that is associated with some measure of entropy or variance that we broaden after each detected change. For example, the Gamma distribution can broaden its scale, and for the categorical distribution, we can increase its entropy/temperature. More intricate (e.g. multimodal) possible alternatives for posterior approximation are also possible, for example, Gaussian mixtures.%

\section{Conclusions}
\label{sec:conclusions}
We introduced variational beam search: an approximate inference algorithm for Bayesian online learning on non-stationary data with irregular changes. Our approach  mediates the tradeoff between a model's ability to memorize past data while still being able to adapt to change. It is based on a Bayesian treatment of a given model's parameters and aimed at tuning them towards the most recent data batch while exploiting prior knowledge from previous batches. To this end, we introduced a sequence of a discrete change variables whose value controlled the way we regularized the model. For no detected change, we regularized the new learning task towards the previously learned solution; for a detected change, we broadened the prior to give room for new data evidence. This procedure is combined with beam search over the discrete change variables. In different experiments, we showed that our proposed model (1) achieved lower error in supervised setups, and (2) revealed a more interpretable and compressible latent structure in unsupervised experiments.  

\paragraph{Broader Impacts.} As with many machine learning algorithms, there is a danger that more automation could potentially result in unemployment. Yet, more autonomous adaptation to changes will enhance the safety and robustness of deployed machine learning systems, such as self-driving cars. 

\section*{Acknowledgements}

We gratefully acknowledge extensive contributions from Robert Bamler (previously UCI, now at the University of T{\"u}bingen), which were indispensable to this work. 

This material is based upon work supported by the National Science Foundation under the CAREER award 2047418 and grant numbers 1633631, 1928718, 2003237, and 2007719; by the National Science Foundation Graduate Research Fellowship under grant number DGE-1839285; by the Defense Advanced Research Projects Agency (DARPA) under Contract No. HR001120C0021; by an Intel grant; and by grants from Qualcomm. Any opinion, findings, and conclusions or recommendations expressed in this material are those of the authors and do not necessarily reflect the views of the National Science Foundation, nor do they reflect the views of DARPA. Additional revenues potentially related to this work include: research funding from NSF, NIH, NIST, PCORI, and SAP; fellowship funding from HPI; consulting income from Amazon.com.

\bibliography{sections/refs.bib}

\newpage

\appendix

\section{Structured Variational Inference}
\label{sec:app-structvi}
According to the main paper, we consider the generative model $p(\x,\z,s_t|\vx_{1:t-1}, s_{1:t-1})=p(s_t)p(\z|s_t; \btau_t)p(\x|\z)$ at time step $t$, where the dependence on $\vx_{1:t-1}, s_{1:t-1}$ is contained in $\btau_t$. Upon observing the data $\x$, both $\z$ and $s_t$ are inferred. However, exact inference is not available due to the intractability of the marginal likelihood $p(\vx_t|s_{1:t}, \vx_{1:t-1})$. To tackle this, we utilize structured variational inference for both the latent variables $\z$ and the Bernoulli change variable $s_t$. To this end, we define the joint variational distribution $q(\z, s_t) = q(s_t|s_{1:t-1})q(\z|s_{1:t})$ as in the main paper. For notational simplicity, we omit the dependence on $s_{1:t-1}$. Then the updating procedure for $q(s_t)$ and $q(\z|s_t)$ is obtained by maximizing the ELBO ${\cal L}(q)$:
\begin{align*}
    &~q^*(\z, s_t) = \argmax_{q(\z, s_t)\in Q} {\cal L}(q), \\
    {\cal L}(q) &:= \E_q[\log p(\x, \z, s_t;\btau_t) - \log q(\z, s_t)].
\end{align*}
Given the generative models, we can further expand ${\cal L}(q)$ to simplify the optimization:
\begin{align}
\label{eq:elbo}
    {\cal L}(q) = &~\E_{q(s_t)q(\z|s_t)} [\log p(s_t) + \log p(\z|s_t;\btau_t) + \log p(\x|\z) - \log q(s_t) - \log q(\z|s_t)] \nonumber \\
     = &~\E_{q(s_t)} [\log p(s_t) - \log q(s_t) + \E_{q(\z|s_t)} [\log p(\z|s_t;\btau_t) + \log p(\x|\z) - \log q(\z|s_t)]] \nonumber \\
     = &~\E_{q(s_t)} [\log p(s_t) - \log q(s_t) + \E_{q(\z|s_t)} [\log p(\x|\z)] - \textrm{KL}(q(\z|s_t)||p(\z|s_t;\btau_t))] \nonumber \\
     = &~\E_{q(s_t)} [\log p(s_t) - \log q(s_t) + {\cal L}(q|s_t)]
\end{align}
where the second step pushes inside the expectation with respect to $q(\z|s_t)$, the third step re-orders the terms, and the final step utilizes the definition of CELBO (Eq.~\ref{eq:cond-elbo} in the main paper). 

Maximizing Eq.~\ref{eq:elbo} therefore implies a two-step optimization: first maximize the CELBO ${\cal L}(q|s_t)$ to find the optimal $q^*(\z|s_t=1)$ and $q^*(\z|s_t=0)$, 
then compute the Bernoulli distribution $q^*(s_t)$ by maximizing ${\cal L}(q)$ while the CELBOs ${\cal L}(q^*|s_t)$ are fixed. 

While $q^*(\z|s_t)$ typically needs to be inferred by black box variational inference~\citep{ranganath2014black, kingma2013auto, zhang2018advances}, the optimal $q^*(s_t)$ has a closed-form solution and bears resemblance to the exact inference counterpart (Eq.~\ref{eq:exact-inference} in the main paper). To see this, we assume ${\cal L}(q^*|s_t)$ are given and $q(s_t)$ is parameterized by $m \in \Rset$ (for the Bernoulli distribution). Rewriting Eq.~\ref{eq:elbo} gives
\begin{align*}
    {\cal L}(q) = &~ m (\log p(s_t=1) - \log m + {\cal L}(q^*|s_t=1))\\ & + (1-m) (\log p(s_t=0) - \log (1-m) + {\cal L}(q^*|s_t=0))
\end{align*}
which is concave since the second derivative is negative. Thus taking the derivative and setting it to zero leads to the optimal solution of
\begin{align*}
    \log \frac{m}{1-m} = &~\log p(s_t=1) - \log p(s_t=0) + {\cal L}(q^*|s_t=1)) - {\cal L}(q^*|s_t=0)), \\
    m = &~\sigma({\cal L}(q^*|s_t=1)) - {\cal L}(q^*|s_t=0)) + \xi_0),
\end{align*}
which attains the closed-form solution as stated in Eq.~\ref{eq:update_s} in the main paper without temperature $T$.

\section{Additive vs.~Multiplicative Broadening}\label{sec:app-broadening}
There are several possible choices for defining an informative prior corresponding to $s_t=1$. In latent time series models, such as Kalman filters~\citep{kalman1960,bamler2017dynamic}, it is common to define a linear transition model $\vz_t=A\vz_{t-1}+\beps_t$ where $\vz_{t-1}\sim{\cal N}(\vz_{t-1}; \bmu_{t-1}, \Sigma_{t-1})$ and $\beps_t\sim{\cal N}(\beps; \mathbf{0}, \Sigma_n)$. Propagating the posterior at time $t-1$ to the prior at time $t$ results in $\vz_t\sim{\cal N}(\vz_t; A\bmu_{t-1}, A\Sigma_{t-1}A^\top + \Sigma_n)$. To simplify the discussion, we set $A=I$  and $\Sigma_n=\sigma_n^2I$; the same argument also applies for the more general case. Adding a constant noise $\beps_t$ results in adding the variance of all variables with a constant $\sigma_n^2$. 
We thus call this convolution scheme \emph{additive broadening}. 
The problem with such a choice, however, is that the associated information loss is not homogeneously distributed: $\sigma_n^2$ ignores the uncertainty in $\z$, and dimensions of $\z$ with low posterior uncertainty lose more information relative to dimensions of $\z$ that are already uncertain. We found that this scheme deteriorates the learning signal. 

We therefore consider \emph{multiplicative broadening} (or \emph{relative broadening} since the associated information loss depends on the original variance) as \emph{tempering} described in the main paper, resulting in $p(\vz_t|s_t, \btau_t) \propto  p(\vz_{t-1}|\vx_{1:t-1}, s_{1:t-1})^{\beta}$ for $\beta>0$. For a Gaussian distribution, the resulting variance scales the original variance with $\frac{1}{\beta}$.
In practice, we found relative or multiplicative broadening to perform much better and robustly than additive posterior broadening. Since tempering broadens the posterior non-locally, this scheme does not possess a continuous latent time series interpretation~\footnote{This means that it is impossible to specify a conditional distribution $p(\bz_t|\bz_{t-1})$ that corresponds to relative broadening.}. 

\section{Details on ``Shy'' Variational Greedy Search and Variational Beam Search}
\label{sec:app-beamsearch}
\begin{figure*}
    \centering
    \includegraphics[width=.9\linewidth]{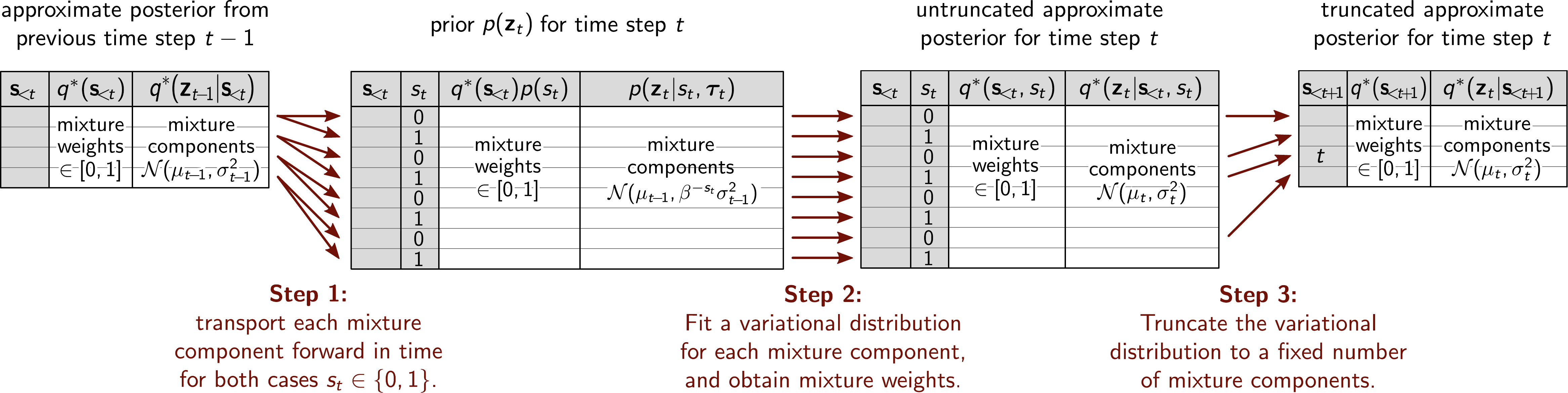}
    \caption{Conditional probability table of variational beam search}
    \label{fig:beam-search-table}
\end{figure*}

\begin{algorithm}[t]
\caption{\label{algo:vbs} Variational Beam Search}
\begin{algorithmic}[1]
\REQUIRE task set $\{\x\}_1^T$; beam size $K$; prior log-odds $\xi_0$; conditional ELBO temperature $T$
\ENSURE approximate posterior distributions $\{q^*(s_{1:t}), q^*(\z|s_{1:t})\}_1^T$
\STATE $q^*(\vz_1)=\argmax \E_q[\log p(\vx_1|\bz_1)] - \text{KL}(q(\vz_1)||p(\vz_1))$;
\STATE $q^*(s_1=0):=1;\; q^*(\bz_1|s_1):=q^*(\bz_1);\; \Bset=\{s_1=0\}$;
\FOR{$t=1,\cdots,T$}
  \STATE $\Bset' = \{\}$
  \FOR{each hypothesis $\bs_{<t}\in\Bset$}
    \STATE $p(s_t = 1) := \sigma(\xi_0)$ for random variable $s_t \in \{0, 1\}$ %
    \STATE $\Bset' := \Bset' \cup \{(\bs_{<t}, s_t=0), (\bs_{<t}, s_t=1)\}$;
    \STATE compute the task $t$'s prior $p(\z|s_t, \btau_t)$ (Eq.~\ref{eq:cond-broaden});
    \STATE perform structured variational inference (Eq.~\ref{eq:cond-elbo} and Eq.~\ref{eq:update_s}) given observation $\x$, resulting in $q^*(s_t,\z|{\bs}_{<t})=q^*(s_t|{\bs}_{<t})q^*(\bz_t|{\bs}_{<t+1})$ where $q^*(\bz_t|{\bs}_{<t+1}$ is stored as output $q^*(\bz_t|s_{1:t})$;
    \STATE approximate new hypotheses' posterior probability $p(s_{1:t}|\bx_{1:t})\approx q^*(\bs_{<t}, s_t) = q^*(\bs_{<t})q^*(s_t|{\bs}_{<t})$;
  \ENDFOR
  \STATE $\Bset:=\;$ \verb#diverse_truncation#$(\Bset', q^*(\bs_{<t}, s_t))$;
  \STATE normalize $q^*(\bs_{<t}, s_t)$ where $(\bs_{<t}, s_t)\in\Bset$;
\ENDFOR
\end{algorithmic}
\end{algorithm}

\paragraph{``Shy'' Variational Greedy Search.} As illustrated in Fig.~\ref{fig:filtering} in the main text, one obtains better interpretation if one outputs the variational parameters $\mu_t$ and $\sigma_t$ at the end of a segment of constant $\z$. More precisely, when the algorithm detects a change point $s_t=1$, it outputs the variational parameters $\mu_{t-1}$ and $\sigma_{t-1}$ from just before the detected change point $t$. These parameters define a variational distribution that has been fitted, in an iterative way, to all data points since the preceding detected change point. We call this the “shy” variant of the variational greedy search algorithm, because this variant quietly iterates over the data and only outputs a new fit when it is as certain about it as it will ever be. The red lines and regions in Fig.~\ref{fig:filtering}~(a) in the main paper illustrate means and standard deviations outputted by the ``shy'' variant of variational greedy search.

We applied this ``Shy'' variant to our illustrative example (Section~\ref{sec:exp-toydata}) and unsupervised learning experiments (Section~\ref{sec:exp-embeddings}).

\paragraph{Variational Beam Search.}
As follows, we present a more detailed explanation of the variational beam search procedure mentioned in Section~\ref{sec:variational-infer} of the main paper. 
Our beam search procedure defines an effective way to search for potential hypotheses with regards to sequences of inferred change points. 
The procedure is completely defined by detailing three sequential steps, that when executed, take a set of hypotheses found at time step $t-1$ and transform them into the resulting set of likely hypotheses for time step $t$ that have appropriately accounted for the new data seen at $t$. 
The red arrows in Figure~\ref{fig:beam-search-table} illustrate these three steps for beam search with a beam size of $K=4$.

In Figure~\ref{fig:beam-search-table}, each of the three steps maps a table of considered histories to a new table. Each table defines a mixture of Gaussian distributions where each mixture component corresponds to a different history and is represented by a different row in the table. 
We start on the left with the (truncated) variational distribution $q^*(\bz_{t-1})$ from the previous time step, which is a mixture over $K=4$ Gaussian distributions. 
Each mixture component (row in the table) is labeled by a 0-1 vector ${\bs}_{<t}\equiv(s_1,\cdots,s_{t-1})$ of the change variable values according to that history. 
Each mixture component ${\bs}_{<t}$ further has a mixture weight $q^*({\bs}_{<t})\in[0,1]$, a mean, and a standard deviation.

We then obtain a prior for time step $t$ by transporting each mixture component of $q^*(\bz_{t-1})$ forward in time via the broadening functional (“Step 1” in the above figure). 
The prior $p(\z)$ (second table in the figure) is a mixture of $2K$ Gaussian distributions because each previous history splits into two new ones for the two potential cases $s_t\in\{0,1\}$. 
The label for each mixture component (table row) is a new vector $({\bs}_{<t},s_t)$ or ${\bs}_{<t+1}$, appending $s_t$ to the tail of ${\bs}_{<t}$.

“Step 2” in the above figure takes the data $\x$ and fits a variational distribution $q^*(\z)$ that is also a mixture of $2K$ Gaussian distributions. 
To learn the variational distribution, we (i) numerically fit each mixture component $q(\z|{\bs}_{<t},s_t)$ individually, using the corresponding mixture component of $p(\z)$ as the prior; (ii) evaluate (or estimate) the CELBO of each fitted mixture component, conditioned on $({\bs}_{<t},s_t)$; (iii) compute the approximate posterior probability $q^*(s_t|{\bs}_{<t})$ of each mixture component, in the presence of the CELBOs; and (iv) obtain the mixture weight equal to the posterior probability over $({\bs}_{<t},s_t)$, i.e., $p(s_{1:t}|\bx_{1:t})$, best approximated by $q^*({\bs}_{<t})q^*(s_t|{\bs}_{<t})$.

“Step 3” in the above figure truncates the variational distribution by discarding $K$ of the $2K$ mixture components. 
The truncation scheme can be either the “vanilla” beam search or diversified beam search outlined in the main paper. 
The truncated variational distribution $q_t(\z)$ is again a mixture of only $K$ Gaussian distributions, and it can thus be used for subsequent update steps, i.e., from $t$ to $t+1$.

The pseudocode is listed in Algo~\ref{algo:vbs}.

\section{Online Bayesian Linear Regression with Variational Beam Search}
\label{sec:app-linreg}

This section will derive the analytical solution of online updates for both Bayesian linear regression and the probability of change points. We consider Gaussian prior distributions for weights. The online update of the posterior distribution is straightforward in the natural parameter space, where the update is analytic given the sufficient statistics of the observations. If we further allow to temper the weights' distributions with a fixed temperature $\beta$, then this corresponds to multiplying each element in the precision matrix by $\beta$. We applied this algorithm to the linear regression experiments in Section~\ref{sec:exp-linreg}. For unified names, we still use the word ``variational'' even though the solutions are analytical.

\subsection{Variational Continual Learning for Online Linear Regression}

Let's start with assuming a generative model at time $t$:
\begin{align}
    \vtheta& \sim\gN(\vmu,\Sigma), \nonumber\\
    y_t& = \vtheta^\top \vx_t + \eps, \quad \eps\sim\gN(0,\sigma_n^2)\label{eq:pred-dist},
\end{align}
and the noise $\eps$ is constant over time.

The posterior distribution of $\vtheta$ is of interest, which is Gaussian distributed since both the likelihood and prior are Gaussian. To get an online recursion for $\vtheta$'s posterior distribution over time, we consider the natural parameterization. The prior distribution under this parameterization is 
\begin{align*}
    p(\vtheta)& = \frac{1}{Z} \exp \left( -\frac{(\vtheta - \vmu)^\top \Sigma^{-1} (\vtheta - \vmu)}{2} \right) \\
    & = \frac{1}{Z} \exp \left( -\frac{1}{2}\vtheta^\top \Sigma^{-1} \vtheta + \vtheta^\top \Sigma^{-1} \vmu \right) \\
    & = \frac{1}{Z} \exp \left( -\frac{1}{2}\vtheta^\top \Lambda \vtheta + \vtheta^\top \veta \right)
\end{align*}
where $\Lambda=\Sigma^{-1}, \veta=\Sigma^{-1} \vmu$ are the natural parameters and the terms unrelated to $\vtheta$ are absorbed into the normalizer $Z$. 

Following the same parameterization, the posterior distribution can be written
\begin{align*}
    p(\vtheta|\vx_t, y_t) & \propto p(\vtheta) p(y_t|\vx_t, \vtheta) \\
    & = \frac{1}{Z} \exp \left( -\frac{1}{2}\vtheta^\top \Lambda \vtheta + \vtheta^\top \veta - \frac{1}{2}\sigma_n^{-2} \vtheta^\top (\vx_t\vx_t^\top) \vtheta + \sigma_n^{-2} y_t\vtheta^\top\vx_t \right) \\
    & = \frac{1}{Z} \exp \left( -\frac{1}{2}\vtheta^\top (\Lambda + \sigma_n^{-2}\vx_t\vx_t^\top) \vtheta + \vtheta^\top (\veta + \sigma_n^{-2} y_t \vx_t) \right). 
\end{align*}
Thus we get the recursion over the natural parameters
\begin{align*}
    \Lambda' & = \Lambda + \sigma_n^{-2}\vx_t\vx_t^\top, \\
    \veta' & = \veta + \sigma_n^{-2} y_t \vx_t,
\end{align*}
from which the posterior mean and covariance can be solved.

\subsection{Prediction and Marginal Likelihood}
We can get the posterior predictive distribution for a new input $\vx_*$ through inspecting Eq.~\ref{eq:pred-dist} and utilizing the linear properties of Gaussian. Assuming the generative model as specified above, we replace $\vx_t$ with $\vx_*$ in Eq.~\ref{eq:pred-dist}. Since $\vtheta$ is Gaussian distributed, by its linear property, $\vx_*^\top \vtheta$ conforms to $\gN(\vx_*^\top \vtheta; \vx_*^\top \vmu, \vx_*^\top \Sigma \vx_*)$. Then the addition of two independent Gaussian results in $y_* \sim \gN(y_*; \vx_*^\top \vmu, \sigma_n^2 + \vx_*^\top \Sigma \vx_*)$.

The marginal likelihood shares this same form with the posterior predictive distribution, with a potentially different pair of sample $(\vx, y)$. To see this, given a prior distribution $\vtheta\sim\gN(\vtheta; \vmu, \Sigma)$, then the marginal likelihood of $y|\vx$ is
\begin{align}
    p(y|\vx; \vmu, \Sigma, \sigma_n) & = \int p(y|\vx, \vtheta) p(\vtheta; \vmu, \Sigma) d\vtheta \nonumber\\
    & = \gN(y; \vx^\top \vmu, \sigma_n^2 + \vx^\top \Sigma \vx) \label{eq:marginal-likelihood}
\end{align}
with $\sigma_n^2$ being the noise variance. Note that in variational inference with an intractable marginal likelihood (not like the linear regression here), this is the approximated objective (Evidence Lower Bound (ELBO), indeed) we aim to maximize.

\paragraph{Computation of the Covariance Matrix}
Since we parameterize the precision matrix instead of the covariance matrix, the variance of the new test sample requires to take the inverse of the precision matrix. In order to do this, we employ the eigendecomposition of the precision matrix and re-assemble to the covariance matrix through inverting the eigenvalues. A better approach is to apply the Sherman-Morrison formula for the rank one update\footnote{\url{https://en.wikipedia.org/wiki/Sherman\%E2\%80\%93Morrison_formula}}, which can reduce the computation from $O(n^3)$ to $O(n^2)$.

\paragraph{Logistic Normal Model}
If we are modeling the log-odds by a Bayesian linear regression, then we need to map the log-odds to the interval $[0,1]$ by the sigmoid function, to make it a valid probability. Specifically, suppose $a=\gN(a; \mu_a, \sigma_a^2)$ and $y=\sigma(a)$ (note we abuse $\sigma$ by variances and functions, but it is clear from the context and the subscripts) where $\sigma(\cdot)$ is a logistic sigmoid function. Then $y$ has a \href{https://en.wikipedia.org/wiki/Logit-normal_distribution}{logistic normal distribution}. Given $p(y)$, we can make decisions for the value of $y$. There are three details that worth noting. First is from the non-linear mapping of $\sigma(\cdot)$. One special property of $p(y)$ is that $p(y)$ can be bimodal if the base variance or $\sigma_a$ is large. A consequence is that the mode of $p(a)$ does not necessarily correspond to $p(y)$'s mode and $\E[y]\neq\sigma(\E[a])$. Second is for the binary classification: the decision boundary of $y$, i.e., 0.5, is consistent with the one of $x$, i.e., 0, for decisions either by $\E[y]$ or by $\E[a]$. See \citet{rusmassen2005gaussian} (Section~3.4) and \citet{bishop1995neural} (Section 10.3). Third, if our loss function for decision making is the absolute error, then the best prediction is the median of $y=\sigma(\hat a)$~\citep{friedman2001elements} where $\hat a$ is the median of $a$. This follows from the monotonicity of $\sigma(\cdot)$ that does not change the order statistics.

\subsection{Inference over the Change Variable}
To infer the posterior distribution of $s_t$ given observations $(\vx_{1:t}, y_{1:t})$, we apply Bayes' theorem to infer the posterior log-odds as in the main paper
\begin{equation*}
    \log\left(\frac{p(s_t=1|\vx_{1:t}, y_{1:t}, s_{1:t-1})}{p(s_t=0|\vx_{1:t}, y_{1:t}, s_{1:t-1})}\right) = \frac{\log(p(\vx_t|\vx_{1:t-1}, y_{1:t}, s_t=1, s_{1:t-1}))}{\log(p(\vx_t|\vx_{1:t-1}, y_{1:t}, s_t=0), s_{1:t-1})} + \xi_0
\end{equation*}
where $p(\vx_t|\vx_{1:t-1}, y_{1:t}, s_t)$ is exactly Eq.~\ref{eq:marginal-likelihood} but has different parameter values dictated by $s_t$ and $\beta$.

In the next part, we show the resulting distribution of the broadening operation.

\subsubsection{Tempering a Multivariate Gaussian}
We will show the tempering operation of a multivariate Gaussian corresponds to multiplying each element in the precision matrix by the fixed temperature, a simple form in the natural space.

Suppose we allow to temper / broaden the $\vtheta$'s multivariate Gaussian distribution before the next time step, accommodating more evidence. Let the broadening constant or temperature be $\beta\in(0, 1]$. We derive how $\beta$ affects the multivariate Gaussian precision.

Write the tempering explicitly,
\begin{align*}
    p(\vtheta)^\beta & = \frac{1}{Z} \exp \left( -\frac{1}{2}\beta(\vtheta - \vmu)^\top \Lambda (\vtheta - \vmu) \right) \\
    & = \frac{1}{Z} \exp \left( -\frac{1}{2}(\vtheta - \vmu)^\top \Lambda_\beta (\vtheta - \vmu) \right)
\end{align*}
among which we are interested in the relationship between $\Lambda_\beta$ and $\Lambda$ and $\beta$. To this end, re-write the quadratic form in the summation
\begin{align*}
    &~\beta(\vtheta - \vmu)^\top \Lambda (\vtheta - \vmu) \\
    = &~\sum_{i,j}\beta\Lambda_{ij}(\vtheta_i - \vmu_i)(\vtheta_j - \vmu_j) \\
    = &~\sum_{i,j}\Lambda_{\beta,ij}(\vtheta_i - \vmu_i)(\vtheta_j - \vmu_j)
\end{align*}
where we can identify an element-wise relation: for all possible $i,j$
\begin{equation*}
    \Lambda_{\beta,ij}=\beta\Lambda_{ij}.
\end{equation*}

\subsubsection{Prediction}
As above, we are interested in the posterior predictive distribution for a new test sample $(\vx_*, y_*)$ after absorbing evidence. Let's denote the parameters of $\vtheta$'s posterior distributions by $\vmu_{s_{1:t}}$ and $\Sigma_{s_{1:t}}$, where the dependence over $s_{1:t}$ is made explicit. We then make posterior predictions with each component \[p(y_*|\vx_*, \vx, y, s_{1:t})={\cal N}(y_*; \vx_*^\top \vmu_{s_{1:t}}, \sigma_n^2 + \vx_*^\top \Sigma_{s_{1:t}} \vx_*).\]

\section{Visualization of Catastrophic Remembering Effects}
\label{sec:app-remember}

\begin{figure}
    \centering
     \includegraphics[width=\columnwidth]{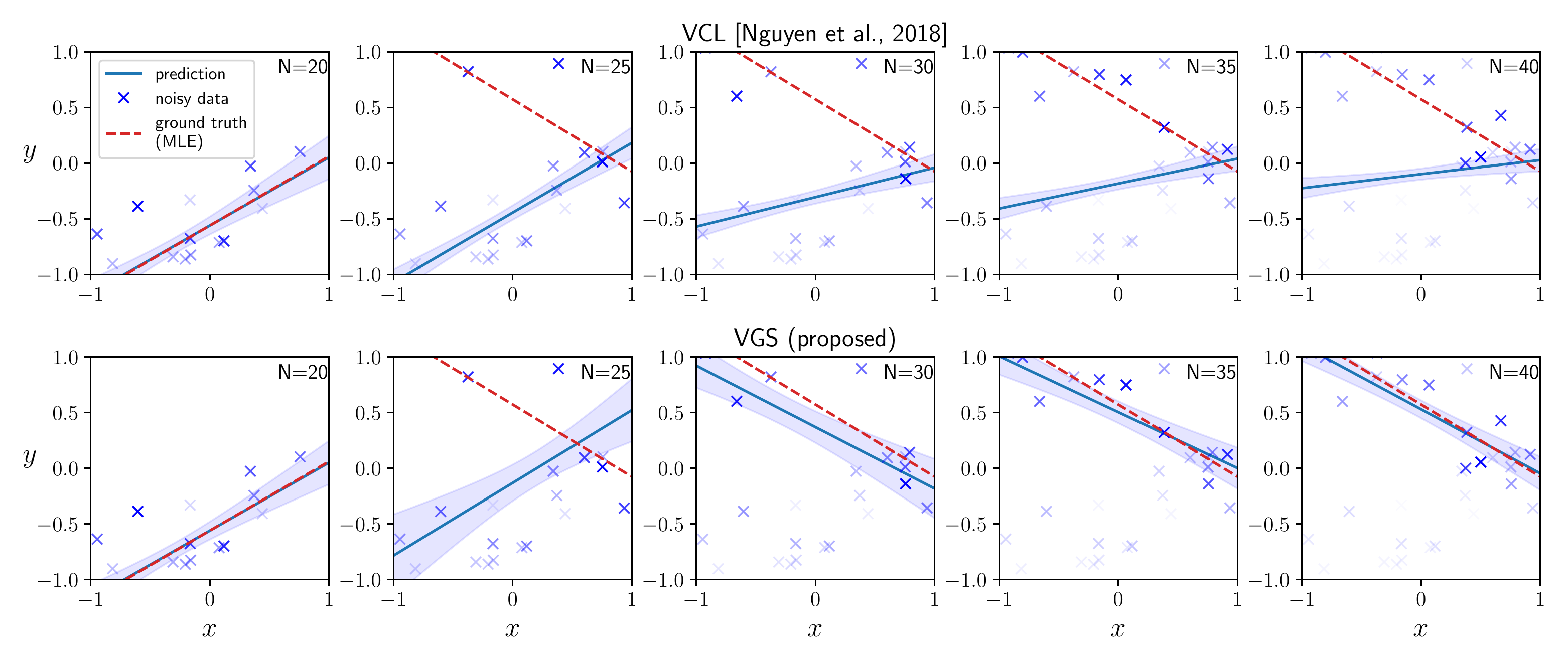}
     \caption{1D online Bayesian linear regression with distribution shift. More recent samples are colored darker. Due to the catastrophic remembering, VCL fails to adapt to new observations.
     }
    \label{fig:remember}
\end{figure}

In order to demonstrate the effect of catastrophic remembering, we consider a simple linear regression model. We will see that, when the data distribution changes, a Bayesian online learning framework becomes quickly overconfident and unable to adjust to the changing data distribution. On the other hand, with tempering, variational greedy search (VGS) can partially forget the previous knowledge and then adapt to the shifted distribution.

\paragraph{Data Generating Process} We generated the samples by the following generative model: 
\begin{align*}
    x &\sim \text{Unif}(-1, 1), \\
    y &\sim \gN(f(x), 0.1^2)
\end{align*}
where $f(x)$ equals $f_1(x)=0.7x-0.5$ or $f_2(x)=-0.7x+0.5$. In this experiments, we sampled the first 20 points from $f_1$ and the remaining 20 points from $f_2$.

\paragraph{Model Parameters} We applied the Bayes updates mentioned in Section~\ref{sec:app-linreg} to do inference over the slope and intercept. We set the initial priors of the weights to be standard Gaussian and the observation noise $\sigma_n^2$ to be the true scale, 0.1. This setting is enough for VCL.

For VGS, we set the same noise variance $\sigma_n^2=0.1$. For the method-specific parameters, we set $\xi_0=\log(0.35/(1-0.35))$ and $\beta=1/3.5$.

We plotted the noise-free posterior predictive distribution for both VCL and VGS. That is, let $f_*(x)$ be the fitted function, we plotted $p(f_*(x_t)|x_t, \gD_{1:t-1})=\int p(f_*(x_t)|x_t, \vw, \gD_{1:t-1})p(\vw|\gD_{1:t-1}) d\vw$ where $\gD_{1:t-1}$ is the observed samples so far.

\paragraph{Results} We first visualized the catastrophic remembering effect through a 1D online Bayesian linear regression task where a distribution shift occurred unknown to the regressor (Fig.~\ref{fig:remember}). In this setup, noisy data were sampled from two ground truth functions $f_1(x)=0.7x-0.5$ and $f_2(x)=-0.7x+0.5$, where, with constant additive noise, the first 20 samples came from $f_1$ and the remaining 20 samples were from $f_2$. The observed sample is presented one by one to the regressor. Before the regression starts, the weights (slope and intercept) were initialized to be standard Gaussian distributions. We experimented two different online regression methods, original online Bayesian linear regression (VCL~\citep{nguyen2017variational}) and the proposed variational greedy search (VGS). In Fig.~\ref{fig:remember}, to show a practical surrogate for the ground truth, we plotted the maximum likelihood estimation (MLE) for each function given the observations. The blue line and the shaded area correspond to the mean and standard deviation of the posterior predictive distribution after observing $N$ samples. As shown in Fig.~\ref{fig:remember}, both VCL (top) and VGS (bottom) faithfully fit to $f_1$ after observing the first 20 samples. However, when another 20 new observations are sampled from $f_2$, VCL shows catastrophic remembering of $f_1$ and cannot adapt to $f_2$. VGS, on the other hand, tempers the prior distribution automatically and succeeds in adaptation. 

\section{NBAPlayer: Change Point Detection Comparisons}
\label{sec:app-nbaplayer}
\begin{figure}
    \centering
     \includegraphics[width=0.38\columnwidth]{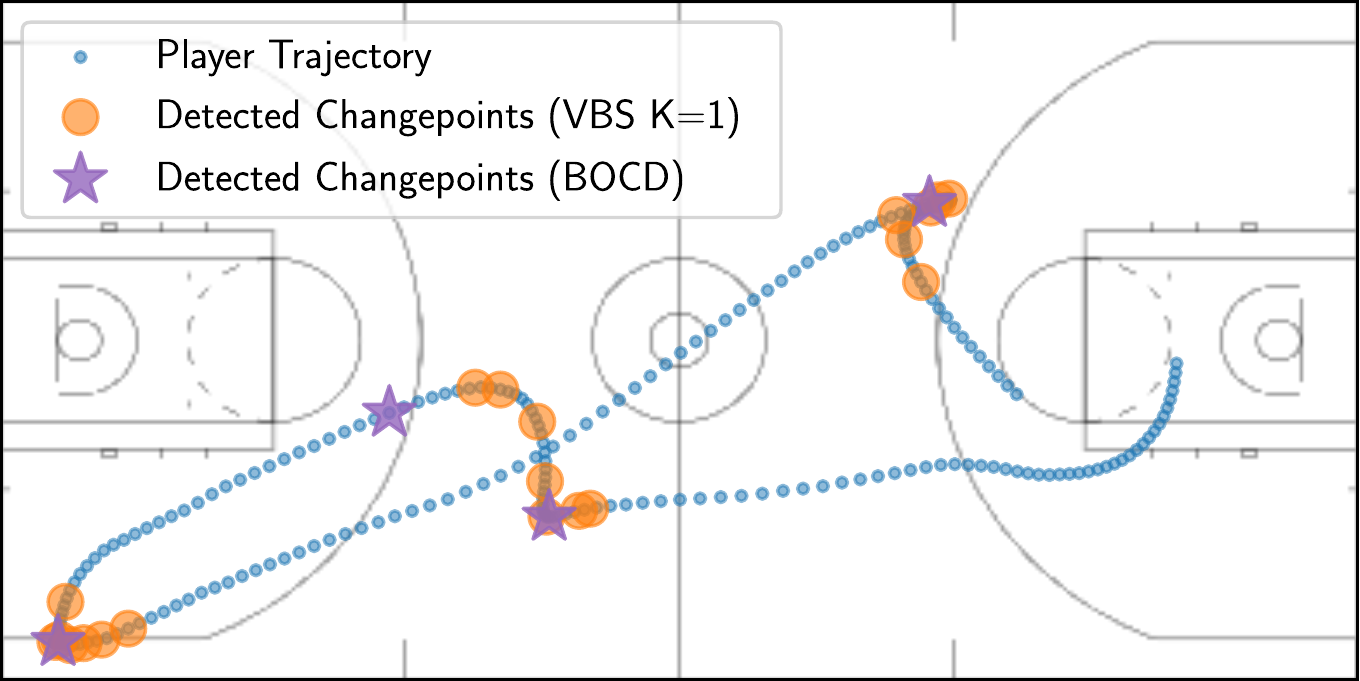}
     \includegraphics[width=0.55\columnwidth]{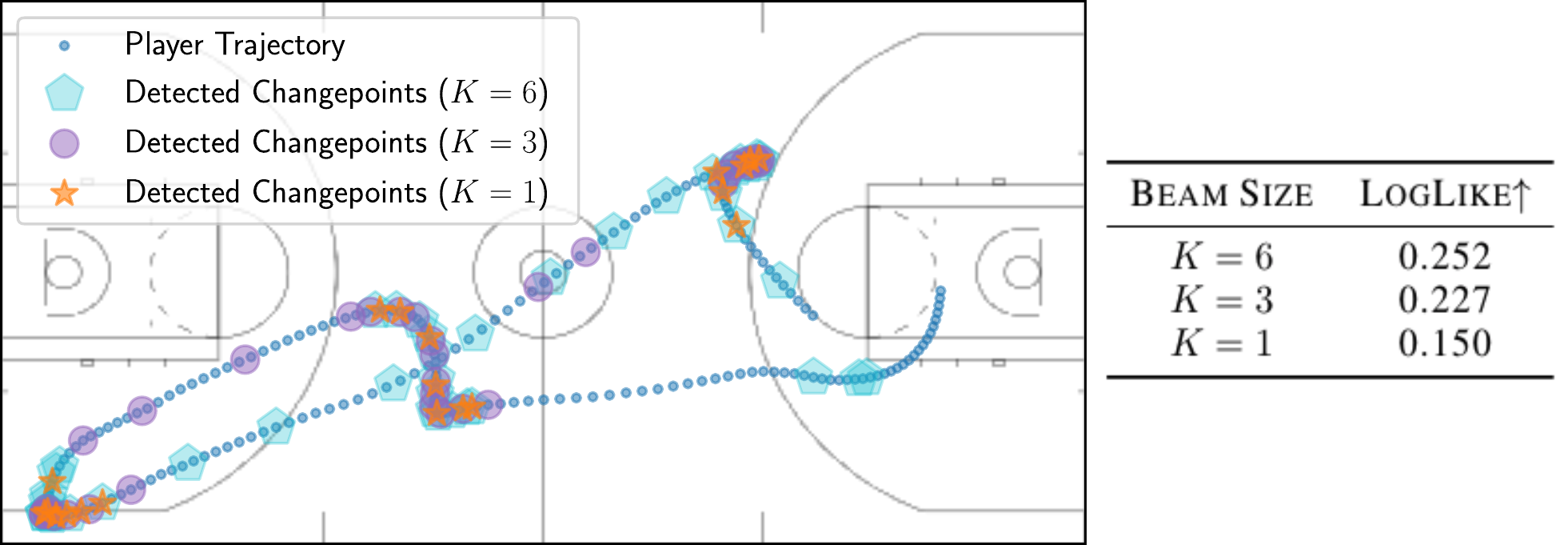}  
     \caption{Changepoints in Basketball player movement tracking. (\textbf{Left plot}) Comparisons between VBS and BOCD~\citep{adams2007bayesian}. 
     While BOCD detect change points at sparse, abrupt changes, VBS detects the changes at smooth, gradual changes. (\textbf{Right plot}) Ablation study over beam size $K$ for VBS while fixing other parameters. As we increase the beam size, qualitatively different change points are detected and the predictive likelihood improves.
     }
    \label{fig:player-tracking}
\end{figure}

\paragraph{VBS vs. BOCD} We investigated the changepoint detection characteristics of our proposed methods and compared against the BOCD baseline in Fig.~\ref{fig:player-tracking} (left). On the shown example trajectory, BOCD detects abrupt change points, corresponding to different plays, a similar phenomenon observed by \citep{harrison2020continuous}. However, we argue that it is insufficient and late to identify a player's strategy purpose -- it only triggers an alarm after a new play starts. VBS, on the other hand, characterizes the transition phases between plays, triggers an early alarm before the next play starts. It also shows the difference between BOCD and VBS in changepoint detection: while BOCD only detects abrupt changes, VBS detects gradual changes as well. 

\paragraph{Practical Considerations of VBS and BOCD} Using our variational inference extensions of BOCD, we can overcome the inference difficulty of non-conjugate models. But, considering practical issues, VBS is better in that the detected change points are easy to read from the binary change variable values $s_{1:t}$ and free from post-processing -- a procedure that BOCD has to exercise. BOCD often outputs a sequence of run lengths either online or offline, among which the change points do not always correspond to the time when the most probable run length becomes one. Instead, the run length oftentimes is larger than one when change point happens. Then people have to inspect the run lengths and set a subjective threshold to determine when a change point occurs, which is not data-driven and may incur undesirable detection. For example, in our basketball player tracking experiments, we set the threshold of change points to be 50; VBS, on the other hand, is free from this post-process thresholding and provides multiple plausible, completely data-driven hypotheses of change points.

\paragraph{Ablation Study of Beam Size} The right plot and the table in Fig~\ref{fig:player-tracking} shows, on the example trajectory, the detected change points and the average log-likelihood as the beam size $K$ changes. When $K=1$, VBS characterizes the trajectory where the velocity direction changes; when $K=3$ or $6$, it seems that some parts where the velocity value changes are detected. We also observed that the average predictive log-likelihood improves as $K$ increases.

\section{Experiment Details and Results}
\label{sec:app-expdetails}

In this section, we provide the unstated details of the experiments mentioned in the main paper. These details include but are not limited to hardware infrastructure used to experiment, physical running time, hyperparameter searching, data generating process, evaluation metric, additional results, empirical limitations, and so on. The subsection order corresponds to the experiments order in the main paper. We first provide some limitations of our methods during experiments.

\paragraph{Limitations} Our algorithm is theoretically sound. The generality and flexibility renders a great performance in experiments, however, at the expense of taking more time to search the hyperparameters in a relatively large space. Specifically, there are two hyperparameters to tune: $\xi_0$ and $\beta$. The grid search over these two hyperparameters could be slow. When we further take into account the beam size $K$, it adds more burden in parameter searching. But we give a reference scope to the tuning region where the search should perform. Oftentimes, we use the same parameters across beam sizes.

\subsection{An Illustrative Example}
\label{sec:app-toy}

\paragraph{Data Generating Process}
To generate Figure~\ref{fig:toydata} in the main paper, we used a step-wise function as ground truth, where the step size was 1 and two step positions were chosen randomly. We sampled 30 equally-spaced points with time spacing 1. To get noisy observations, Gaussian noise with standard deviation 0.5 was added to the points.

\paragraph{Model Parameters}
In this simple one-dimensional model, we used absolute broadening with a Gaussian transition kernel $K(\z, \z')={\cal N}(\z-\z', D\Delta t)$ where $D=1.0$ and $\Delta t = 1$. The inference is thus tractable because $p(\z|s_t)$ is conditional conjugate to $p(\x|\z, s_t)$ (and both are Gaussian distributed). We set the prior log-odds $\epsilon_0$ to $\log\frac{p(s_t=1)}{p(s_t=0)}$, where $p(s_t=1)=0.1$. We used beam size 2 to do the inference.

\subsection{Bayesian Linear Regression Experiments}
\label{sec:app-linregexp}

We performed all linear regression experiments on a laptop with 2.6 GHz Intel Core i5 CPU. All models on SensorDrift, Elec2, and NBAPlayer dataset finished running within 5 minutes. Running time on Malware dataset varied: VCL, BF, and Independent task are within 10 minutes; VGS takes about two and half hours; VBS (K=3) takes about six hours; VBS (K=6) takes about 12 hours. BOCD takes similar time with VBS. The main difference between VCL's computation cost and VBS's computation cost lies in the necessarity of inverting the precision matrix into covariance matrix. However, the matrix inverse computation in VBS and BOCD can be substantially reduced from $O(n^3)$ to $O(n^2)$ by the recursion of Sherman–Morrison formula; we will implement this in the future.

\begin{figure}
    \begin{center}
       \includegraphics[width=0.3\columnwidth]{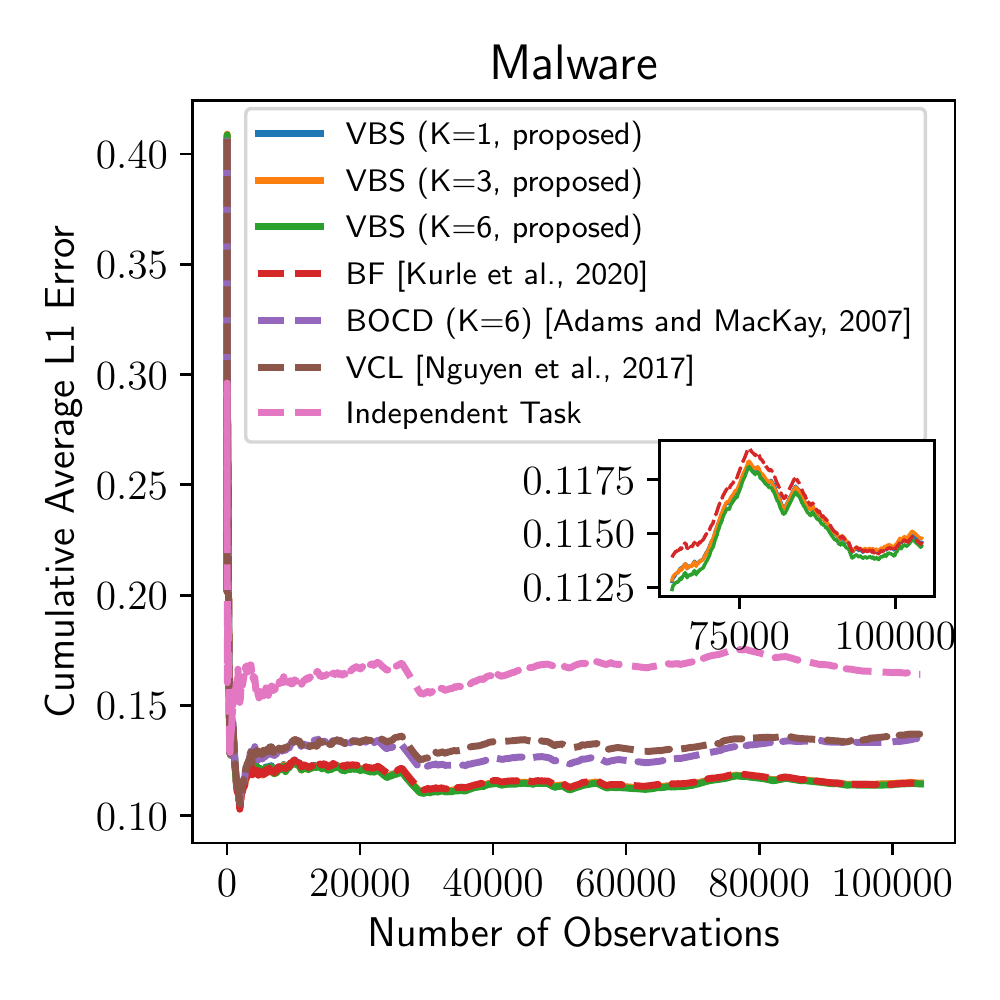}
       \includegraphics[width=0.3\columnwidth]{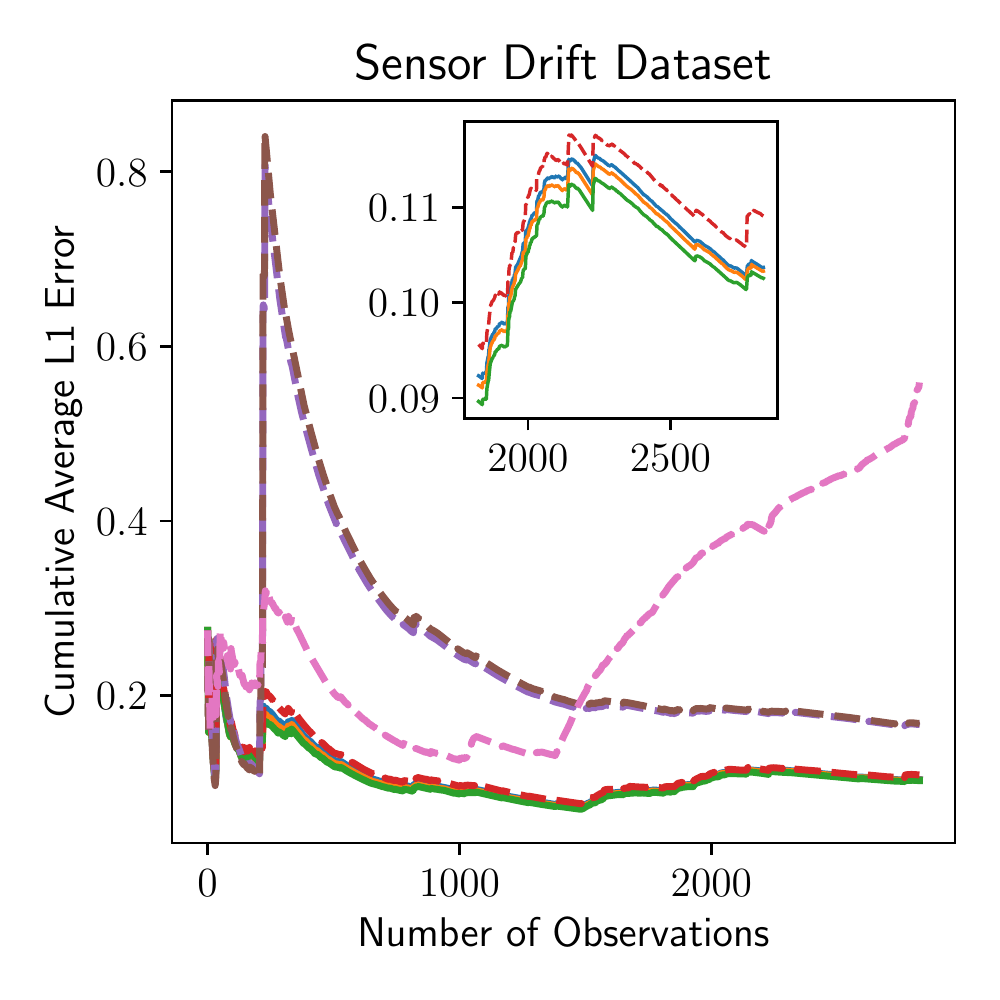}
       \includegraphics[width=0.3\columnwidth]{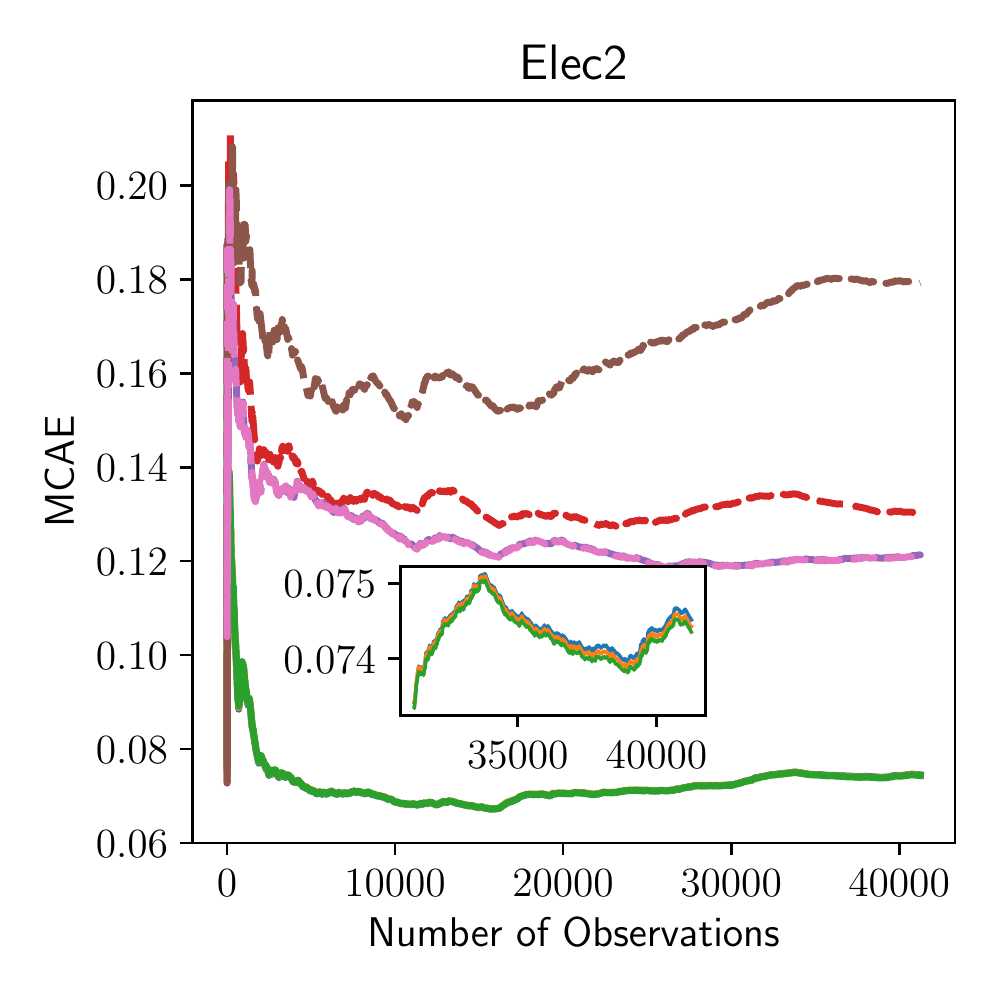}
       \caption{One-step ahead performance of the online malware detection experiments. Proposed methods outperform the baseline.
       } 
    \end{center}
    \label{fig:malware}
\end{figure}

\paragraph{Problem Definitions} We considered both classification experiments (Malware, Elec2) and regression experiments (SensorDrift, NBAPlayer). The classification datasets have real-value probabilities as targets, permitting to perform regression in log-odds space. 

\paragraph{Setup and Evaluation} We defined each task to consist of a single observation. Models made predictions on the next observation immediately after finishing learning current task. Models were then evaluated with one-step-ahead absolute error\footnote{in probability space for classification tasks; in data space for regression tasks. With the exception of NBAPlayer dataset, we evaluated models with predictive log probability $\frac{1}{t}\sum_{i=2}^t \log p(y_i|y_{1:i-1}, x_{1:i})$.}, which is then used to compute the mean cumulative absolute error (MCAE) at time $t$:  
$\frac{1}{t}\sum_{i=1}^t |y^*_i - y_i|$ where $y^*$ is the predicted value and $y_i$ is the ground truth. We further approximated the Gaussian posterior distribution by a point mass centered around the mode. It should be noticed that for linear regression, the Laplace Propagation has the same mode as Variational Continual Learning, and the independent task has the same mode as its Bayesian counterpart.

\paragraph{Results} We reported the result of the dominant hypothesis of VBS with large beam size. Fig.~\ref{fig:malware} shows MCAE over time for the first three datasets. Our methods remain lower prediction error of all time while baselines are subject to strong fluctuations or inability to adapt to distribution shifts. Another observation is that VBS with different beam sizes performed similarly in this case.

\subsubsection{Baseline Hyperparameters}
\paragraph{BOCD} We only keep the top three or six most probable run length after each time step. 

We tuned the hyperparameter $\lambda$ in the hazard function, or the transition probability. 
$\lambda^{-1}$ is searched in $\{0.1, 0.2, 0.3, 0.4, 0.5, 0.6, 0.7, 0.8, 0.9, 0.99\}$.

Malware selects $\lambda^{-1}=0.3$; Elec2 selects $\lambda^{-1}=0.9$; SensorDrift selects $\lambda^{-1}=0.6$; NBAPlayer selects $\lambda^{-1}=0.99$.

\paragraph{BF} We implemented Bayesian Forgetting according to [Kurle et al., 2020]. 

We tuned the hyperparameter $\beta$ as the forgetting rate such that $p(z_t|D_{t-1}) \propto p_0(z_t)^{1-\beta}q_{t-1}(z_t|D_{t-1})^\beta$ where $0<\beta<1$. 
$\beta$ is searched in $\{0.001, 0.01, 0.1, 0.2, 0.3, 0.4, 0.5, 0.6, 0.7, 0.8, 0.9, 0.95, 0.97, 0.98, 0.99, 0.995, 0.999\}$.

Malware selects $\beta=0.999$; Elec2 selects $\beta=0.98$; SensorDrift selects $\beta=0.9$; NBAPlayer selects $\beta=0.9$.

\subsubsection[Malware]{Malware\footnote{\url{https://archive.ics.uci.edu/ml/datasets/Dynamic+Features+of+VirusShare+Executables}}}

\paragraph{Dataset} There are 107856 programs collected from 2010.11 to 2014.7 in the monthly order. Each program has 482 counting features and a real-valued probability $p\in[0,1]$ of being malware. This ground truth probability is the proportion of 52 antivirus solutions that label malware. We used the first-month data (2010.11) as the validation dataset and the remaining data as the test dataset. To enable analytic online update, we cast the binary classification problem in the log-odds space and performed Bayesian linear regression. We filled log-odds boundary values to be $-5$ and $4$, corresponding to probability $0$ and $1$, respectively. Our methods achieved comparable results reported in \citep{huynh2017new} on this dataset.

\paragraph{Hyperparameters} We searched the hyperparameters $\sigma_n^2, \xi_0=\log \frac{p_0}{1-p_0}$, and $\beta$ using the validation set. Specifically, we extensively searched $\sigma_n^2\in\{0.1, 0.2, \cdots, 0.9, 1, 2, \cdots, 10, 20, \cdots, 100\}$, $p_0\in\{0.5\}$, $\beta^{-1}\in\{1.01, 1.05, 1.1, 1.2, 1.5, 2, 5\}$. On most values the optimization landscape is monotonic and thus the search quickly converges around a local optimizer. Within the local optimizer, we performed the grid search, which focused on $\beta\in[1.05, 1.2]$.

We found all methods favored $\sigma_n^2=40$. And for VGS and VBS, the uninformative prior of the change variable $p_0=0.5$ was already a suitable one. VGS selected $\beta^{-1}=1.2$, VBS (K=3) selected $\beta^{-1}=1.07$, and VBS (K=6) selected $\beta^{-1}=1.05$. Although searched $\beta^{-1}$ varies for different beam size, the performance of different beam size in this case, based on our experience, is insensitive to the varying $\beta$.

\subsubsection[SensorDrift]{SensorDrift\footnote{\url{http://archive.ics.uci.edu/ml/datasets/Gas+Sensor+Array+Drift+Dataset+at+Different+Concentrations}}}

\paragraph{Dataset} We focused on one kind of gas, \emph{Acetaldehyde}, retrieved from the original gas sensor drift dataset \citep{vergara2012chemical}, which spans 36 months. We formulated an regression problem of predicting the gas concentration level given all the other features. The dataset contains 128 features and 2926 samples in total. We used the first batch data (in the original dataset) as the validation set and the others as the test set. Since the scales of the features vary greatly, we scaled each feature with the sample statistics computed from the validation set, leading to zero mean and unit standard deviation for each feature.

\paragraph{Hyperparameters} We found that using a Bayesian forgetting module (instead of tempered posterior module mentioned in the main paper), which corresponds to $s_t=1$, works better for this dataset. Since we scaled the dataset, we therefore set the hyperparameter $\sigma_n^2=1$ for all methods. We searched $\xi_0=\log \frac{p_0}{1-p_0}$ and $\beta$ using the validation set. Specifically, we did the grid search for the prior probability of change point $p\in\{0.501, 0.503, 0.505, 0.507, 0.509, 0.511, 521, 0.55, 0.6, 0.7, 0.8, 0.9\}$ and the temperature $\beta\in\{0.5, 0.6, 0.7, 0.8, 0.9\}$. 
The search procedure selects $p=0.507$ and $\beta=0.7$ for all beam size $K$.

\subsubsection[Elec2]{Elec2\footnote{\url{https://www.openml.org/d/151}}}

\paragraph{Dataset} The dataset contains the electricity price over three years of two Australian states, New South Wales and Victoria \citep{harries1999splice}. While the original problem was 0-1 binary classification, we re-produced the targets with real-value probabilities since all necessary information forming the original target is contained in the dataset. Specifically, we re-defined the target to be the probability of the price in New South Wales increasing relative to the price of the last 24 hours. Then we performed linear regression in the log-odds space. We filled log-odds boundary values to be $-4$ and $4$, corresponding to probability $0$ and $1$, respectively. After removing the first 48 samples (for which we cannot re-produce the targets), we had 45263 samples, and each sample comprised 14 features. The first 4000 samples were used for validation while the others were used for test. 

\paragraph{Hyperparameters} We searched the hyperparameters $\sigma_n^2, \xi_0=\log \frac{p_0}{1-p_0}$, and $\beta$ using the validation set. Specifically, we extensively searched $\sigma_n^2\in\{0.01, 0.02, \cdots, 0.1, 0.2, \cdots, 1, 2, \cdots, 10, 20, \cdots, 100\}$, $p_0\in\{0.5\}$, $\beta^{-1}\in\{1.05, 1.1, 1.2, 1.5, 2, 5\}$. 

VCL favored $\sigma_n^2=0.01$, and we set this value for all other methods. VGS selected $\beta^{-1}=1.2$. VBS (K=3) and VBS (K=6) inherited the same $\beta$ value from VGS.

\subsubsection[NBAPlayer]{NBAPlayer\footnote{\url{https://github.com/linouk23/NBA-Player-Movements}}}

\paragraph{Dataset} The original dataset contains part of the game logs of 2015-2016 NBA season in json files. The log records each on-court player's position (in a 2D space) at a rate of 25 Hz. We pre-processed the logs and randomly extracted ten movement trajectories for training set and another ten trajectories for test set. For an instance of the trajectory, we selected Wesley Matthews's trajectory at the 292th event in the game of Los Angeles Clippers vs. Dallas Mavericks on Nov 11, 2015. The trajectories vary in length and correspond to players' strategic movement. After extracting the trajectories, we fix the data set and then evaluate all methods with it. Specifically, we regress the current position on the immediately previous position--modeling the player's velocity. 

\paragraph{Hyperparameters} We searched the hyperparameters $\sigma_n^2, \xi_0=\log \frac{p_0}{1-p_0}$, and $\beta$ using the training set. Specifically, we searched $\sigma_n^2\in\{0.001, 0.01, 0.1, 0.5, 1.0, 10., 100.\}$ and $p\in\{0.501, 0.503, 0.505, 0.507, 0.509, 0.511, 521, 0.55, 0.6, 0.7, 0.8, 0.9\}$ and $\beta\in\{0.5, 0.6, 0.7, 0.8, 0.9\}$. 

VCL favored $\sigma_n^2=0.1$, and we set this value for all other methods. VBS (K=1, VGS) selected $\beta=0.5$ and $p=0.513$. VBS (K=3) selected $p=0.507$ and $\beta=0.6$.  VBS (K=6) selected $p=0.505$ and $\beta=0.7$. In generating the plots on the example trajectory, we used $p=0.507$ with varying $\beta$ and varying beam size $K$.

\subsection{Bayesian Deep Learning Experiments}
\label{sec:app-bdl}

We performed the Bayesian Deep Learning experiments on a server with Intel(R) Xeon(R) Gold 5218 CPU @ 2.30GHz and Nvidia TITAN RTX GPUs. Regarding the running time, VCL, and Independent task (Bayes) takes five hours to finish training; Independent task takes three hours; VGS takes two GPUs and five hours; VBS (K=3) takes six GPUs and five hours; VBS (K=6) takes six GPUs and 10 hours. When utilizing multiple GPUs, we implemented task multiprocessing with process-based parallelism.

\begin{figure}
    \centering
    \includegraphics[width=\textwidth]{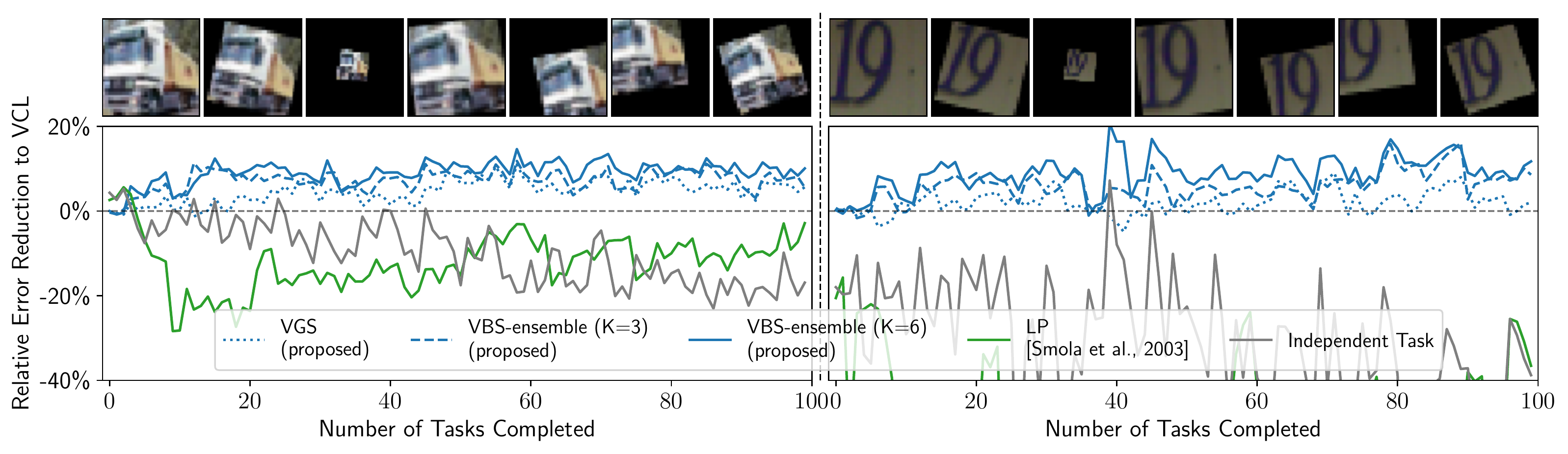}
    \caption{(\textbf{Bottom}) Running test performance of our proposed VBS and VGS algorithms compared to various baselines on transformed CIFAR-10 (left) and SVHN (right). (\textbf{Top}) Examples of transformations that we used for introducing covariate shifts.}
    \label{fig-run-bnn}
\end{figure}

\begin{table*}
\caption{Convolution Neural Network Architecture}
\label{table:archit-cnn}
\begin{center}
\begin{small}
\begin{sc}
\begin{tabular}{lcccccc}
\toprule
layer & filter size & filters & stride & activation & dropout \\
\midrule
Convolutional  & $3\times3$ & 32 & 1 & ReLU & \\
Convolutional  & $3\times3$ & 32 & 1 & ReLU & \\
MaxPooling  & $2\times2$ &  & 2 & & 0.2 \\
Convolutional  & $3\times3$ & 64 & 1 & ReLU & \\
Convolutional  & $3\times3$ & 64 & 1 & ReLU & \\
MaxPooling  & $2\times2$ &  & 2 & & 0.2 \\
FullyConnected  & & 10 & & Softmax & \\
\bottomrule
\end{tabular}
\end{sc}
\end{small}
\end{center}
\end{table*}

\begin{table*}
\caption{Hyerparameters of Bayesian Deep Learning Models for CIFAR-10}
\label{table:hyperparam-cifar10}
\begin{center}
\begin{small}
\begin{sc}
\begin{tabular}{lccccccc}
\toprule
model & learning rate & batch size & number of epochs & $\beta$ & $\xi_0$ or $\lambda^{-1}$ & $T$ \\
\midrule
LP & 0.001 & 64 & 150 & N/A & N/A & N/A\\
BOCD   & 0.0005 & 64 & 150 & N/A & 0.3 & 20000\\
BF   & 0.0005 & 64 & 150 & 0.9 & N/A & 20000\\
VCL   & 0.0005 & 64 & 150  & N/A & N/A & N/A\\
VBS   & 0.0005 & 64 & 150 & 2/3 & 0 & 20000\\
\bottomrule
\end{tabular}
\end{sc}
\end{small}
\end{center}
\end{table*}

\begin{table*}
\caption{Hyperparameters of Bayesian Deep Learning Models for SVHN.}
\label{table:hyperparam-svhn}
\begin{center}
\begin{small}
\begin{sc}
\begin{tabular}{lccccccc}
\toprule
model & learning rate & batch size & number of epochs & $\beta$ & $\xi_0$ or $\lambda^{-1}$ & $T$ \\
\midrule
LP  & 0.001 & 64 & 150 & N/A & N/A & N/A\\
BOCD   & 0.00025 & 64 & 150 & N/A & 0.3 & 20000\\
BF   & 0.00025 & 64 & 150 & 0.9 & N/A & 20000\\
VCL    & 0.00025 & 64 & 150  & N/A & N/A & N/A\\
VBS     & 0.00025 & 64 & 150 & 2/3 & 0 & 20000\\
\bottomrule
\end{tabular}
\end{sc}
\end{small}
\end{center}
\end{table*}

\paragraph{Datasets with Covariate shifts}
We used two standard datasets for image classification: CIFAR-10 \citep{krizhevsky2009learning} and SVHN \citep{netzer2011reading}. We adopted the original training set and used the first 5000 images in the original test set as the validation set and the others as the test set. We further split the training set into batches (or tasks in the continual learning literature) for online learning, each batch consisting of a third of the full data. Each transformation (either rotation, translation, or scaling) is generated from a fixed, predefined distribution (see below for  \textbf{Transformations}) as covariate shifts.
Changes are introduced every three tasks, where the total number of tasks was 100. 

\paragraph{Transformations}
We used Albumentations~\citep{info11020125} to implement the transformations as covariate shifts. As stated in the main paper, the transformation involved rotation, scaling, and translation. Each transformation factor followed a fixed distribution: rotation degree conformed to $\cN(0, 10^2)$; scaling limit conformed to $\cN(0, 0.3^2)$; and the magnitude of vertical and horizontal translation limit conformed to $\text{Beta}(1, 10)$, and the sampled magnitude is then rendered positive or negative with equal probability. The final scaling and translation factor should be the corresponding sampled limit plus 1, respectively.

\paragraph{Architectures and Protocol}
All Bayesian and non-Bayesian methods use the same neural network architecture. We used a truncated version of the VGG convolutional neural network (in Table~\ref{table:archit-cnn})
on both datasets. We confirmed that our architecture achieved  similar performance on CIFAR10 compared to the results reported by \citet{zenke2017continual} and \citet{lopez2017gradient} in a similar setting. %
We implemented the Bayesian models using \href{https://www.tensorflow.org/probability/overview}{TensorFlow Probability} and the non-Bayesian counterpart (namely Laplace Propagation) using \href{https://www.tensorflow.org/versions/r1.15/api_docs/python/tf/keras}{TensorFlow Keras}. Every bias term in all the models were treated deterministically and were not affected by any regularization. 

We initialize each algorithm by training the model on the full, untransformed dataset. 
During every new task, all algorithms are trained until convergence.

\paragraph{Tempered Conditional ELBO}
In the presence of massive observations and a large neural network, posterior distributions of change variables usually have very low entropy because of the very large magnitude of the difference between conditional ELBOs as in Eq.~\ref{eq:update_s}. 
Therefore change variables become over confident about the switch-state decisions. 
The situation gets even more severe in beam search settings where almost all probability mass is centered around the most likely hypothesis while the other hypotheses get little probability and thereby will not take effect in predictions.
A possible solution is to temper the conditional ELBO (or the marginal likelihood) and introduce more uncertainty into the change variables. 
To this end, we divide the conditional ELBO by the number of observations. It is equivalent to set $T=20000$ in Eq.~\ref{eq:update_s}. 
This practice renders every hypothesis effective in beam search setting.

\paragraph{Hyperparameters, Initialization, and Model Training}
The hyperparameters used across all of the models for the different datasets are listed in Tables~\ref{table:hyperparam-cifar10}~and~\ref{table:hyperparam-svhn}. Regarding the model-specific parameters, we set $\xi_0$ to 0 for both datasets and searched $\beta$ in the values $\{5/6, 2/3, 1/2, 1/4\}$ on a validation set. We used the first 5000 images in the original test set as the validation set, and the others as the test set. We found that $\beta=2/3$ performs relatively well for both data sets. Optimization parameters, including learning rate, batch size, and number of epochs, were selected to have the best validation performance of the classifier on one independent task. To estimate the change variable $s_t$'s variational parameter, we approximated the conditional ELBOs~\ref{eq:cond-elbo} by averaging 10000 Monte Carlo samples.

As outlined in the main paper, we initialized each algorithm by training the model on the full, untransformed dataset. The model weights used a standard Gaussian distribution as the prior for this meta-initialization step. 

When optimizing with variational inference, we initialized $q(\z)$ to be a point mass around zero for stability. When performing non-Bayesian optimization, we initialized the weights using Glorot Uniform initializer~\citep{glorot2010understanding}. All bias terms were initialized to be zero.

We performed both the Bayesian and non-Bayesian optimization using ADAM \citep{kingma2014adam}. For additional parameters of the ADAM optimizer, we set $\beta_1 = 0.9$ and $\beta_2=0.999$ for both data sets. For the deep Bayesian models specifically, which include VCL and VBS, we used stochastic black box variational inference \citep{ranganath2014black, kingma2013auto, zhang2018advances}. We also used the Flipout estimator~\citep{wen2018flipout} to reduce variance in the gradient estimator.

\paragraph{Predictive Distributions}
We evaluated the most likely hypothesis' predictive posterior distribution of the test set by the following approximation:
\begin{equation*}
    p(\vy^*|\vx^*, {\cal D}_{1:t}, s_{1:t}) \approx \frac{1}{N} \sum_{n=1}^N p(\vy^*|\vx^*, \bz_{s_{1:t}}^{(n)})
\end{equation*}
where $N$ is the number of Monte Carlo samples from the variational posterior distribution $q^*(\vz_t|s_{1:t})$. 
In our experiments we found $S=10$ to be sufficient. We take $\argmax_{\y} p(\vy^*|\vx^*,, {\cal D}_{1:t})$ to be the predicted class.

LP only used the MAP estimation $\z^*$ to predict the test set: $p(\vy^*|\vx^*, {\cal D}_{1:t}) \approx p(\vy^*|\vx^*,, \z^*)$.

\paragraph{Standard Deviation in the Main Text Table~\ref{table-bdlaccuracy}}
The results in this table were summarized and reported by taking the average over tasks. Each algorithm’s confidence, which is usually evaluated by computing the standard deviation across tasks in stationary environments, now is hard to evaluate due to the non-stationary setup. These temporal image transformations will largely affect the performance, leaving the blindly computed standard deviation meaningless since the standard deviation across all tasks represents both the data transformation variation and the modeling variation. To evaluate the algorithm’s confidence, we proposed a three-stage computation. We first segment the obtained performance based on the image transformations (in our case, we separate the performance sequence every three tasks). Then we compute the standard deviation for every performance segment. Finally, we average these standard deviations across segments as the final one to be reported. In this way, we can better account for the data variation in order to isolate the modeling variation. 

\paragraph{Running Performance.} We also reported the running performance for both our methods and some baselines for each task over time (100 tasks in total) in Fig.~\ref{fig-run-bnn}.  In the bottom panel, to account for varying task difficulties, we show the percentage of the error reduction relative to VCL, a Bayesian online learning baseline. Our proposed approach can achieve 10\% error reduction most of the time on both datasets, showing the adaptation advantage of our approach. The effect of beam search is also evident, with larger beam sizes consistently performing better. The top panel shows some examples of the transformations that we used for introducing covariate shifts manually.

\subsection{Dynamic Word Embeddings Experiments}\label{sec:app-embeddings}

\begin{figure*}
    \centering
    \includegraphics[width=\textwidth]{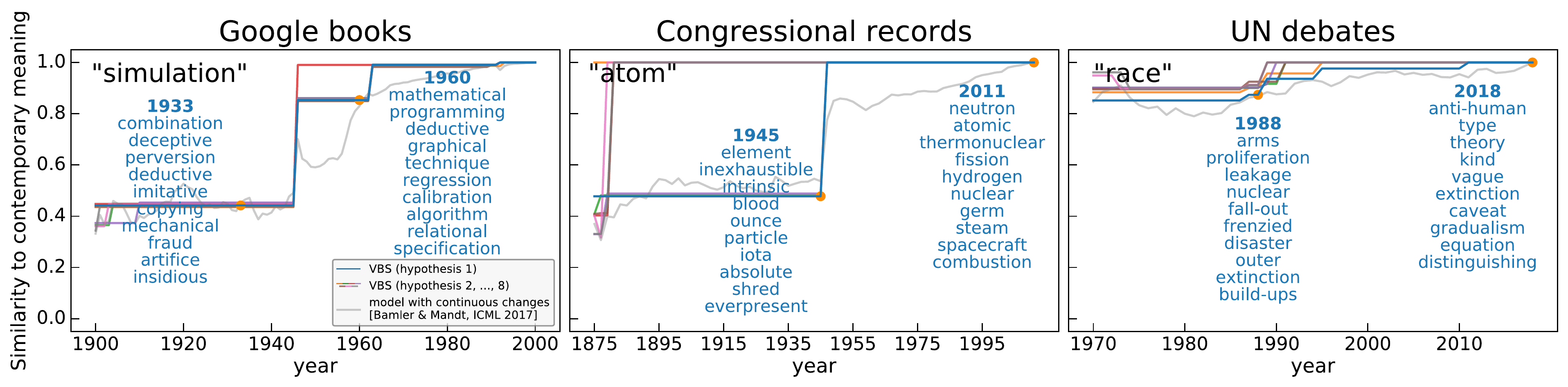}
    \caption{Dynamic Word Embeddings on Google books, Congressional records, and UN debates, trained with VBS~(proposed, colorful) vs. VCL~(grey). In contrast to VCL, VBS reveals sparse, time-localized semantic changes~(see main text).}
    \label{fig:dwe_changepoint_sample}
\end{figure*}
\begin{figure*}[t]
    \centering
    \includegraphics[width=\textwidth]{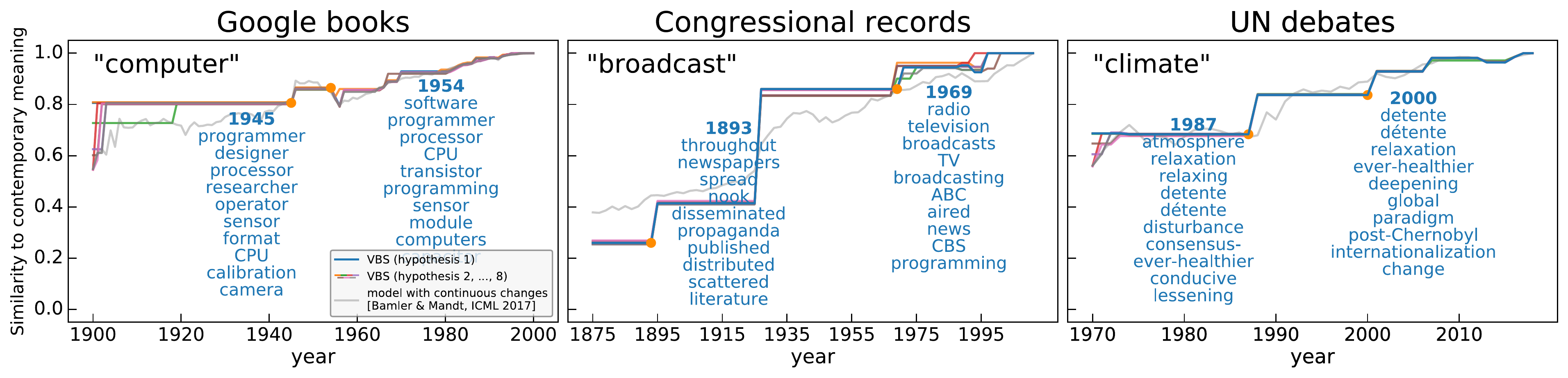}
    \caption{Additional results of Dynamic Word Embeddings on Google books, Congressional records, and UN debates.}
    \label{fig:app-dwe-changepoint}
\end{figure*}

We performed the Dynamic Word Embeddings experiments on a server with Intel(R) Xeon(R) Gold 5218 CPU @ 2.30GHz and Nvidia TITAN RTX GPUs. Regarding the running time, for qualitative experiments, Google Books and Congressional Records take eight GPUs and about 24 hours to finish;  UN Debates take eight GPUs and about 13 hours to finish. For quantitative experiments, since the vocabulary size and latent dimensions are smaller, each model corresponding to a specific $\xi_0$ takes eight GPUs and about one hour to finish. When utilizing multiple GPUs, we implemented task multiprocessing with process-based parallelism.

\begin{table*}
\caption{Hyperparameters of Dynamic Word Embedding Models}
\label{table:hyperparam-dwe}
\begin{center}
\begin{scriptsize}
\begin{sc}
\begin{tabular}{lccccccccc}
\toprule
corpus & vocab & dims & $\beta$ & learning  rate &  epoches & $\xi_0$ & beam size (K) & $T$ \\
\midrule
Google books & 30000 & 100 & 0.5 & 0.01 & 5000 & -10 & 8 & 1 \\
Congressional records & 30000 & 100 & 0.5 & 0.01 & 5000  & -10 & 8 & 1\\
UN debates & 30000 & 20 & 0.25 & 0.01 & 5000 & -1 & 8 & 1\\
\bottomrule
\end{tabular}
\end{sc}
\end{scriptsize}
\end{center}
\end{table*}

\begin{table*}
\caption{$\xi_0$ of Document Dating Tasks}
\label{table:hyperparam-docdating}
\begin{center}
\begin{small}
\begin{sc}
\begin{tabular}{lc}
\toprule
corpus & $\xi_0$  \\
\midrule
Google books & -1000000, -100000, -5120, -1280, -40 \\
Congressional records & -100000, -1280, -320, -40\\
UN debates & -128, -64, -32, -4\\
\bottomrule
\end{tabular}
\end{sc}
\end{small}
\end{center}
\end{table*}

\paragraph{Data and Preprocessing} We analyzed three large time-stamped text corpora, all of which are available online. Our first dataset is the Google Books corpus \citep{michel2011quantitative} consisting of $n$-grams, which is sufficient for learning word embeddings. We focused on the period from 1900 to 2000. To have an approximately even amount of data per year, we sub-sampled 250M to 300M tokens per year. Second, we used the Congressional Records data set \citep{gentzkow2018congressional}, which has 13M to 52M tokens per two-year period from 1875 to 2011. Third, we used the UN General Debates corpus \citep{DVN/0TJX8Y_2017}, which has about 250k to 450k tokens per year from 1970 to 2018.
For all three corpora, the vocabulary size used was 30000 for qualitative results and 10000 for quantitative results.
We further randomly split the corpus of every time step into training set (90\%) and heldout test set (10\%). All datasets, Congressional Records\footnote{\url{https://data.stanford.edu/congress_text}}, UN General Debates\footnote{\url{https://dataverse.harvard.edu/dataset.xhtml?persistentId=doi:10.7910/DVN/0TJX8Y}}, and Google Books\footnote{\url{http://storage.googleapis.com/books/ngrams/books/datasetsv2.html}} can be downloaded online. 

We tokenized Congressional Records and UN General Debates with pre-trained Punkt tokenizer in NLTK\footnote{\url{https://www.nltk.org/}}. We constructed the co-occurence matrices with a moving window of size 10 centered around each word. Google books are already in Ngram format.

\paragraph{Model Assumptions}
As outlined in the main paper, we analyzed the semantic changes of individual words over time. We augmented the probabilistic models proposed by~\citet{bamler2017dynamic} with our change point driven informative  prior (Eq.~\ref{eq:cond-broaden} in the main paper) to encourage temporal sparsity.
We pre-trained the \textit{context} word embeddings\footnote{We refer readers to \citep{mikolov2013distributed, bamler2017dynamic} for the difference between target and context word embeddings.} using the whole corpus, and kept them constant when updating the \textit{target} word embeddings.
This practice denied possible interference on one target word embedding from the updates of the others. 
If we did not employ this practice, the spike and slab prior on word $i$ would lead to two branches of the “remaining vocabulary” (embeddings of the remaining words in the vocabulary), conditioned either on the spike prior of word $i$ or on the slab prior. This hypothetical situation gets severe when every word in the vocabulary can take two different priors, thus leading to exponential branching of the sequences of inferred change points.
When this interference is allowed, the exponential scaling of hypotheses translates into exponential scaling of possible word embeddings for a single target word, which is not feasible to compute for any meaningful vocabulary sizes and number of time steps. 
To this end, while using a fixed, pre-trained context word embeddings induces a slight drop of predictive performance, the computational efficiency improves tremendously and the model can actually be learned. 

\paragraph{Hyperparameters and Optimization}
Qualitative results in Figure~\ref{fig:dwe_changepoint_sample} in the main paper were generated using the hyperparameters in Table~\ref{table:hyperparam-dwe}. The initial prior distribution used for all latent embedding dimensions was a standard Gaussian distribution. We also initialized all variational distributions with standard Gaussian distributions. For model-specific hyperparameters $\beta$ and $\xi_0$, we first searched the broadening constant $\beta$ to have the desired jump magnitude observed from the semantic trajectories mainly for medium-frequency words. We then tuned the bias term $\xi_0$ to have the desired change frequencies in general. We did the searching for the first several time steps. We performed the optimization using black box variational inference and ADAM. For additional parameters of ADAM optimizer, we set $\beta_1 = 0.9$ and $\beta_2=0.999$ for all three corpora. In this case, we did \textbf{not} temper the conditional ELBO by the number of observations (correspondingly, we set $T=1$ in Eq.~\ref{eq:update_s} in the main paper). 

Quantitative results of VGS in Figure~\ref{fig:toydata} (c) in the main paper were generated by setting a smaller vocabulary size and embedding dimension, 10,000 and 20, respectively for all three corpora. We used an eight-hypothesis (K=8) VBS to perform the experiments. Other hyperparameters were inherited from the qualitative experiments except $\xi_0$, whose values used to form the rate-distortion curve can be found in Table~\ref{table:hyperparam-docdating}. We enhanced beam diversification by dropping the bottom two hypotheses instead of the bottom third hypotheses before ranking. On the other hand, the baseline, “binning”, and had closed-form performance if we assume (i) a uniformly distributed year in which a document query is generated, (ii) “binning” perfectly locates the ground truth episode, and (iii) the dating result is uniformed distributed within the ground truth episode. The $L1$ error associated with “binning” with episode length $L$ is $\E_{t\sim {\cal U}(1,L), t'\sim {\cal U}(1,L)} [|t-t'|]=\frac{L-1}{2}$. By varying $L$, we get binning's rate-distortion curve in Figure~\ref{fig:toydata} (c) in the main paper.

\paragraph{Predictive Distributions}
In the demonstration of the quantitative results, i.e., the document dating experiments, we predicted the year in which each held-out document's word-word co-occurrence statistics $\bx$ have the highest likelihood and measured $L1$ error. To be specific, for a given document in year $t$, we approximated its likelihood under year $t'$ by evaluating $\frac{1}{|V|}\log p(\x|\bz_{t'}^*)$, where $\bz_{t'}^*$ is the mode embedding in year $t'$ and $|V|$ is the vocabulary size. We predicted the year $t^*=\argmax_{t'} \frac{1}{|V|}\log p(\x|\bz_{t'}^*)$. We then measured the $L1$ error by $\frac{1}{T}\sum_i^T |t_i - t_i^*|$ given $T$ truth-prediction pairs.

\subsubsection{Additional Results}
\paragraph{Qualitative Results} 
As outlined in the main paper, our qualitative result shows that the information priors encoded with change point detection is more interpretable and results in more meaningful word semantics than the diffusion prior of \citep{bamler2017dynamic}. Here we provide a more detailed description of the results with more examples. Figure~\ref{fig:dwe_changepoint_sample} shows three selected words (``simulation'', ``atom'', and ``race''--one taken from each corpus) and their nearest neighbors in latent space. As time progresses the nearest neighboring words change, reflecting a semantic change of the words. While the horizontal axis shows the year, the vertical axis shows the cosine distance of the word's embedding vector at the given year to its embedding vector in the last available year.

The plot reveals several interpretable semantic changes of each word captured by VBS. For example, 
as shown by the most likely hypothesis in blue for the Congressional Records data, 
the term “atom” changes its meaning  from ``element'' to ``nuclear'' in 1945--the year when two nuclear bombs were detonated. The word “race” changes from the cold-war era “arms”(-race) to its more prevalent meaning after 1991 when the cold war ended. The word ``simulation'' changes its dominant context from ``deception'' to ``programming'' with the advent of computers. 
The plot also showcases various possible semantic changes of all eight hypotheses, where each hypothesis states various aspects.

Additional qualitative results can be found in Figure~\ref{fig:app-dwe-changepoint}. it, again, reveals interpretable semantic changes of each word: the first change of “computer” happens in 1940s--when modern computers appeared; “broadcast” adopts its major change shortly after the first commercial radio stations were established in 1920; “climate” changes its meaning at the time when Intergovernmental Panel on Climate Change~(IPCC) was set up, and when it released the assessment reports to address the implications and potential risks of climate changes.

\paragraph{Quantitative Results and Baseline}

Figure~\ref{fig:toydata} (c) in the main paper shows the results on the three corpora data, where we plot the document dating error as a function of allowed changes per year. For fewer allowed semantic changes per year, the dating error goes up. Lower curves are better. %

Now we describe how the baseline ``Binning'' was constructed.
We assumed that we had separate word embeddings associated with episodes of $L$ consecutive years. For $T$ years in total, the associated memory requirements would be proportional to $V*T/L$, where $V$ is the vocabulary size. Assuming we could perfectly date the document up to $L$ years results in an average dating error of $\frac{L}{2}$. We then adjusted $L$ to different values and obtained successive points along the "Binning" curve.

\end{document}